\documentclass[10pt,twocolumn,letterpaper]{article}

\usepackage[pagenumbers]{cvpr} % To force page numbers, e.g. for an arXiv version

\usepackage[dvipsnames]{xcolor}

\definecolor{cvprblue}{rgb}{0.21,0.49,0.74}
\usepackage[pagebackref,breaklinks,colorlinks,citecolor=cvprblue]{hyperref}
\usepackage{amsmath,epstopdf,amssymb,color,url,multirow,multicol,verbatim,threeparttable,diagbox,makecell,booktabs,colortbl}
\usepackage[numbers,sort&compress]{natbib} %合并参考文献
\def\paperID{4480} % *** Enter the Paper ID here
\def\confName{CVPR}
\def\confYear{2024}

\usepackage{floatrow}
\newfloatcommand{capbtabbox}{table}[][\FBwidth]

\newcommand{\tft}{\textbf}
\newcommand{\et}{\textit{et al.}}

\newcommand{\cred}{\color{red}}
\newcommand{\cblue}{\color{blue}}

\definecolor{light-gray}{gray}{0.82}

\title{Rethinking the Up-Sampling Operations in CNN-based Generative Network for Generalizable Deepfake Detection}
\author{Chuangchuang Tan$^{1,2}$, Huan Liu$^{1,2}$, Yao Zhao$^{1,2}$\thanks{Corresponding author}, Shikui Wei$^{1,2}$, Guanghua Gu$^{3,4}$, Ping Liu$^{5}$, Yunchao Wei$^{1,2}$ \\
{\small $^1$Institute of Information Science, Beijing Jiaotong University}\\
{\small $^2$Beijing Key Laboratory of Advanced Information Science and Network Technology} \\
{\small $^3$School of Information Science and Engineering, Yanshan University}\\
{\small $^4$Hebei Key Laboratory of Information Transmission and Signal Processing}\\
{\small $^5$Center for Frontier AI Research, IHPC, A*STAR, Singapore}\\
{\tt\small tanchuangchuang@bjtu.edu.cn}
}

\begin{document}
\maketitle
\begin{abstract}

Recently, the proliferation of highly realistic synthetic images, facilitated through a variety of  GANs and Diffusions, has significantly heightened the susceptibility to misuse. While the primary focus of deepfake detection has traditionally centered on the design of detection algorithms, an investigative inquiry into the generator architectures has remained conspicuously absent  in recent years. 
This paper contributes to this lacuna by rethinking the architectures of CNN-based generator, thereby establishing a generalized representation of synthetic artifacts.  Our findings illuminate that the up-sampling operator can, beyond frequency-based artifacts, produce generalized forgery artifacts. In particular, the local interdependence among image pixels caused by upsampling operators is significantly demonstrated in synthetic images generated by GAN or diffusion. Building upon this observation, we introduce the concept of Neighboring Pixel Relationships(NPR) as a means to capture and characterize the generalized structural artifacts stemming from up-sampling operations. A comprehensive analysis is conducted on an open-world dataset, comprising samples generated by \tft{28 distinct generative models}. This analysis culminates in the establishment of a novel state-of-the-art performance, showcasing a remarkable \tft{11.6\%} improvement over existing methods. The code is available at \cblue{\url{https://github.com/chuangchuangtan/NPR-DeepfakeDetection}}.

\end{abstract}    
\section{Introduction}
\label{sec:intro}

 \begin{figure}[ht!]
   \centering
    \includegraphics[scale=0.270]{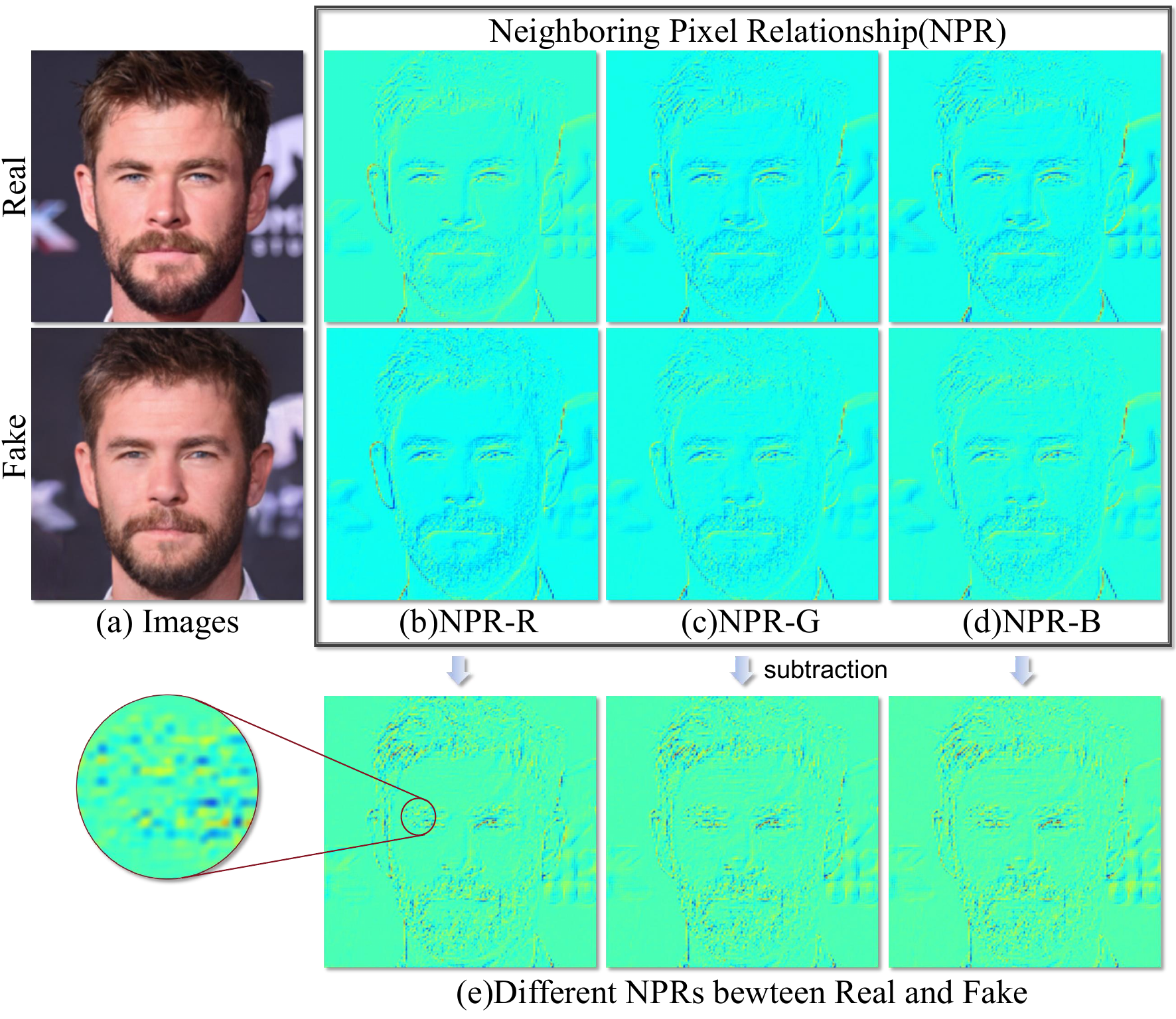}
    \caption{\tft{The visualization of Neighboring Pixel Relationship (NPR) of real image and its inversion \cite{Idinvert}}. To fully understand the NPR, (a) we invert the real image by \cite{Idinvert}, and (b-d) present NPR heatmap for the R, G, B channel of images. In addition, to show that NPR can be used as artifacts representation, (e) the differential NPRs between real and fake images is shown. The proposed NPR Effectively reveals the differences between real and fake images.}
    \label{fig:nprh}
 \end{figure}

% 第一段
% 生成发展很迅速
% 需要伪造

% 通用是必须的 

% 第二段
% 伪造发展了几年，大家都集中在几个方面，但是在通用性/泛化性上仍然缺乏

% 第三段
% in this paper, aiming to extract generalized artifacts, we focus on performing the analysis of generator arturech. 
% the frequency-based methods have proved that 上采样在GAN中很常见/common. 

% 第四段
% It 会产生频域伪影，可以作为检测表示。近期的研究发现频域的泛化性不好。this work rethinking the upsampling layer 
% 介绍生成模型结构，上采样在其中很通用

% 已经有论文搞了，但是有问题

% 上采样如何产生新的伪影                       % 关键在此处   如何用数学公式表示

% 新的伪影如何有效，有什么意义            % 关键在此处

% 金字塔

% 效果

% contribution
% NPR
% NPR-Pyramid or NPR-Patch
% 14%

%%%%%%%%%%%%%%%%%%%%%%%%%%%%%%%%%%%%%%%%%%%%%%%%%%

% 第一段
% 生成发展很迅速
% 需要伪造
With the rapid evolution of image synthetic technologies, such as GAN\cite{goodfellow2014generative, karras2018progressive, karras2019style}, Diffusion\cite{ho2020denoising,rombach2022high}, AI-generated images have reached a level of realism that makes them virtually indistinguishable from authentic images to human observers.
Nevertheless, the misuse of these capabilities poses potential threats in political and economic domains. 
Addressing this issue requires the development of generalizable deepfake detection methods.
In recent years, notable strides \cite{chollet2017xception, Frank, Durall} have been made in forgery detection, particularly in the face forgery detection.

In the realm of deepfake detection, a significant challenge for detectors is to generalize effectively to unseen deepfake sources in real-world scenarios. 
Recent advancements aimed at enhancing this generalization ability include the refinement of detection algorithms \cite{qian2020thinking,jeong2022bihpf}, the augmentation of datasets through the acquisition of more diverse images \cite{jeong2022fingerprintnet,chen2022ost,jeong2022frepgan}, and the development of pre-trained models \cite{Tan2023CVPR,ojha2023towards}. 
Despite these efforts, a conspicuous gap remains in the lack of source-invariant representation exploited from the generator pipeline for forgery image detection. This deficiency leads to failures in detecting unknown forgery domains. Intriguingly, there has been a scarcity of investigative inquiry into generator architectures in recent years.

%There are lack of source-invariant representation from generator pipeline for forgery image detection, resulting in fail to the unknown forgery domains. 
%The investigative inquiry into the generator architectures has remained conspicuously absent in recent years. %so far. 

% 通用是必须的 

% 第二段
% 伪造发展了几年，大家都集中在几个方面，但是在通用性/泛化性上仍然缺乏

% 第三段
% in this paper, aiming to extract generalized artifacts, we focus on performing the analysis of generator architectures. 
% the frequency-based methods have proved that 上采样在GAN中很常见/common.

In addressing this challenge, our work centers on analyzing generator architectures to extract generalized artifact representations.
Previous studies \cite{zhang2019detecting, Frank, Durall} have demonstrated the ubiquity of up-sampling components in common GAN pipelines.
Simultaneously, given the widespread adoption of U-Net in diffusion models, such as DDPM \cite{ho2020denoising}, ADM \cite{dhariwal2021diffusion}, and LDM \cite{rombach2022high}, the up-sampling layer emerges as a crucial module in diffusion models.
The up-sampling cue holds significant potential for advancing generalizable deepfake detection.
Building on these insights, current research delves into the influence of up-sampling across the entire image, developing the frequency spectrum as a representation of up-sampling artifacts.
However, recent findings \cite{jeong2022frepgan} suggest that frequency-based artifacts may not suffice for achieving generalization in detection, given the diverse patterns present in the frequency domain of GANs.

%Aiming to address this problem, this work focuses on performing the analysis of generator architectures to extract generalized artifacts representation. 
%Previous works \cite{zhang2019detecting, Frank, Durall} have shown that the up-sampling component is included in the common GAN pipelines. 
%Meanwhile, due to the U-Net  is adopted widely in diffusion models, such as DDPM \cite{ho2020denoising}, ADM \cite{dhariwal2021diffusion}, LDM \cite{rombach2022high}, the up-sampling layer is also one of the most important modules of diffusion models. 
%The cue of up-sampling have the huge potential of generalizable deepfake detection. 
%Based on this insights, existing works consider the influence of the whole image, and develop the frequency spectrum as the up-sampling artifacts. 

% 已经有论文搞了，但是有问题
%However, some works \cite{jeong2022frepgan} find out that frequency-based artifacts are not up to the generalization detection, since various patterns of frequency domain in GANs.
%We also show the spectrum of different generation sources. 
%There are different patterns of spectrum in GANs or  real images of different classes. 
%The frequency-based artifact representation is not enough generalized to the unseen sources.

In this paper, our focus is on achieving source-invariant forgery detection by rethinking artifacts stemming from the up-sampling component of common generation models. 
Existing works predominantly consider its impact on the entire image in the frequency domain. 
In contrast, our approach involves exploring the trace of the up-sampling layer at the level of local image pixels, providing a more nuanced understanding of its influence.

%In this paper, to achieve source-invariant forgery detection, we rethinking the artifacts from the up-sampling component of common generation models.  
%Existing works only consider its influence on the whole image in the frequency domain.
%In contrast, we explore the trace of up-sampling layer from the local image pixels. % 缺一句

Specifically, in the pipelines of common generation models, up-sampling is employed to transform the low-resolution latent space into high resolution.
Within the scaled feature, local pixels exhibit a strong relationship. For instance, employing nearest neighbor interpolation results in the local $2 \times 2$ pixels sharing the same value.
Subsequent to the up-sampling operation, the scaled features are further processed through convolutional layers to generate images. 
During this process, a relationship is established among local pixels through the combination of the up-sampling operation and the translation invariance of CNN layers. 
This, in turn, manifests as discernible relationships among local pixels in the generated images.
%We leverage this intrinsic relationship induced by the up-sampling operator as source-invariant forgery artifacts. Through experimental evidence, we will demonstrate the presence of these local up-sampling artifacts in both GANs and diffusion models.

%Specifically, in the common generation models pipelines, the up-sampling are used to transform the low resolution latent space to the high resolution. In the scaled feature, the local pixels have a strong relationship. For example, if up-sampling operation is nearest neighbor interpolation, the local $2 \times 2$ pixels would have same value. 
%After up-sampling, the scaled feature are furthre fed into the convolutional layers, utile generate the images. 
%In this processing, strong relationship built in the input by up-sampling operation, and translation Invariance of CNN layer. 
%Those result in the relationship of local pixels shown in the generated images.
%We adopt this relationship caused by the up-sampling operator as source-invariant forgery artifacts.
%We will show through experiments that both GANs and diffusions exhibit the local up-sampling artifacts.

Building upon these insights, we propose a simple but effective artifact representation, termed Neighboring Pixel Relationships (NPR), aimed at achieving generalized deepfake detection. NPR serves as the artifact representation for training the detection model.
The primary innovation of our approach lies in introducing a simple yet versatile artifact representation derived from the common up-sampling component of generation pipelines. In Fig. \ref{fig:nprh}, we showcase NPR heatmaps for a real face and its inversion. Significantly, NPR effectively captures artifacts related to image details such as hair, eyes, and beard. Despite the generator's tendency to enhance details for realism, traces of the up-sampling layer persist in the local image pixels.
%In addition, drawing inspiration from the Laplacian Pyramid, we introduce the Neighboring Pixel Relationships Pyramid (NPRP) to further enhance generalization. NPRP is constructed by aggregating Multi-scale Neighboring Pixel Relationships.

% 新的伪影如何有效，有什么意义            % 关键在此处 逻辑
%Based on this insights, we propose a simple but effective artifact representation, named Neighboring Pixel Relationships (NPR), to achieve generalized deepfake detection. We adopt NPR as the artifact representation to train the detection model.
%The key innovation of our methods lies in present a simple and generalized artifact representation from the common up-sampling component of generation pipelines. 
%The NPR heatmaps of the real face and its inversion are shown in Fig. \ref{fig:nprh}.  
%It can be observed that our NPR can capture the artifacts from image details, such as hair, eye, and beard. 
%Although the generator tend to make the details clearer and more realistic, the traces of up-sampling layer still lie in the local image pixels. 
%In addition, inspired by Laplacian Pyramid, we introduce the Neighboring Pixel Relationships Pyramid (NPRP) to enhance the generalization ability. The NPRP is built by fusing the Multi-scale Neighboring Pixel Relationships. 

% 第四段
% It 会产生频域伪影，可以作为检测表示。近期的研究发现频域的泛化性不好。this work rethinking the upsampling layer 
% 介绍生成模型结构，上采样在其中很通用

% 已经有论文搞了，但是有问题

% 上采样如何产生新的伪影                       % 关键在此处   如何用数学公式表示

% 新的伪影如何有效，有什么意义            % 关键在此处

% 金字塔

% 效果
To comprehensively evaluate the generalization ability of our proposed NPR, we conduct simulations using a vast database of images generated by 28 distinct models~\footnote{ProGAN, StyleGAN, StyleGAN2, BigGAN, CycleGAN, StarGAN, GauGAN, Deepfake, AttGAN, BEGAN, CramerGAN, InfoMaxGAN, MMDGAN, RelGAN, S3GAN, SNGAN, STGAN, DDPM, IDDPM, ADM, LDM, PNDM, VQDiffusion, Glide, Stable Diffusion v1, Stable Diffusion v2,  DALLE, and Midjourney.}.
Our extensive experiments demonstrate the effectiveness and versatility of the artifact representation generated by the NPR across diverse and unseen sources.

% contribution
% NPR
% NPR-Pyramid or NPR-Patch
% 14%
Our paper makes the following contributions:
\begin{itemize}

\item We propose a simple yet effective artifact representation, Neighboring Pixel Relationships (NPR), designed to capture local up-sampling artifacts from image pixels, thereby achieving generalized forgery image detection.
Thanks to the widespread use of up-sampling operations in existing generation models, NPR demonstrates the ability to generalize to unseen sources, covering unknown GAN or Diffusion models.

%\item We extend the Neighboring Pixel Relationships method by introducing Neighboring Pixel Relationships Pyramid(NPRP). This extension combines artifact representations from Multi-scale Neighboring Pixel Relationships, enhancing detection performance by leveraging local up-sampling artifacts with diverse fields to generate more generalized representations.

\item We demonstrate that up-sampling operators can cause generalized forgery artifacts beyond frequency-based artifacts. The trace of the up-sampling layer from local image pixels exhibits more generalization compared to its influence on the whole image in the frequency domain for deepfake detection.

\item Our experiments validate the effectiveness of the proposed NPR, showcasing strong generalization capabilities across 28 different generation models used for forgery image synthesis. 
%This underscores the versatility of NPR in handling various unseen sources.
We observe a remarkable gain of $11.6\%$ compared to existing methods, highlighting the effectiveness and superiority of NPR in real-world scenarios.

\end{itemize}

%\section{Related Work}
%\label{sec:related_work}

\section{Related Work}
%In this section, we provide a brief survey on deepfake detection approaches, which can be classified into two main categories: image-based and frequency-based detection.
In this section, we present a concise survey of deepfake detection approaches, categorizing them into two main groups: image-based and frequency-based detection.
% \Ping{1. The content size of the "Image-based" section appears to be larger than that of the "Frequency-based" section. Please maintain consistency and balance in the document.
% 2. It would be beneficial to avoid writing any paragraphs that are excessively long. Keeping paragraphs concise and focused will enhance the readability and clarity of the document.
% }
% \ccn{I have reduced the Image-based.}
\subsection{Image-based Fake Detection}
Some studies \cite{Deepfake} utilize images as input data to train binary classification models for forgery detection. Rossler et al. \cite{Deepfake} employ images to train a straightforward Xception \cite{chollet2017xception} for detecting fake face images. Other works concentrate on specific regions, such as eyes and lips, to discern fake face media \cite{li2018ictu, haliassos2021lips}. 
Yu \etal \cite{yu2019attributing} and Marra \etal \cite{marra2019gans} extract the unique fingerprints of the GAN model from generated images to perform detection. 
Chai \etal \cite{chai2020makes} use limited receptive fields to find the patches which make images detectable. 
Chai et al. \cite{chai2020makes} employ limited receptive fields to identify patches that render images detectable. Some works enhance the generalization of detectors to unseen sources by diversifying training data through augmentation methods \cite{wang2020cnn, wang2021representative}, adversarial training \cite{chen2022self}, reconstruction techniques \cite{cao2022end, he2021beyond}, fingerprint generators \cite{jeong2022fingerprintnet}, and blending images \cite{shiohara2022detecting}. Additionally,
 Ju et al. \cite{ju2022fusing} integrate global spatial information and local informative features to train a two-branch model.
 The AltFreezing \cite{wang2023altfreezing} adopts both spatial and temporal artifacts to achieve Face Forgery Detection.
 Ojha et al. \cite{ojha2023towards} and Tan et al. \cite{Tan2023CVPR} employ feature maps and gradients, respectively, as general representations.

\subsection{Frequency-based Fake Detection}
Given that GAN architectures heavily rely on up-scaling operations,  some studies \cite{Durall, Frank} delve into the impact of up-sampling across the entire image, developing the frequency spectrum as a representation of up-sampling artifacts. LOG \cite{masi2020two} integrates information from both color and frequency domains to detect manipulated face images and videos. 
F3-Net \cite{qian2020thinking} introduces frequency components partition and the discrepancy of frequency statistics between real and forged images into face forgery detection. Luo et al. \cite{luo2021generalizing} utilize multiple high-frequency features of images to enhance generalization performance. ADD \cite{woo2022add} develops two distillation modules for detecting highly compressed deepfakes, including frequency attention distillation and multi-view attention distillation. BiHPF \cite{jeong2022bihpf} amplifies the magnitudes of artifacts through two high-pass filters. FreGAN \cite{jeong2022frepgan} observes that unique frequency-level artifacts in generated images can lead to overfitting to training sources. Consequently, FreGAN mitigates the impact of frequency-level artifacts through frequency-level perturbation maps.

\section{Methodology}
\label{sec:method}

% 不同上采样的训练数据。
% 需要讨论

Efforts in achieving generalizable forgery detection often aim to develop a detector trained on one or a few sources that can effectively generalize to other sources.% within an open-world setting. 
To accomplish this objective, our work is dedicated to designing a generalizable artifacts representation through an analysis of common up-sampling operations in popular generators. We introduce a form of local up-sampling artifacts, named Neighboring Pixel Relationships (NPR), the details of which are presented in this section. 
%Furthermore, we introduce NPRP as an extension of our approach. 
%The overall architectures of NPR and NPRP are depicted in Fig. \ref{}.

%The generalizable forgery detection tend to obtain a detector trained on one or few sources, which is able to generalize to other sources from open-world scene.   
%In order to achieve this goal,  this work designs a generalizable artifacts representation by analyzing the common up-sampling operations in the popular generator. 
%We propose a local up-sampling artifacts, named Neighboring Pixel Relationships (NPR), and will present its details in this section. 
%Different from the global up-sampling artifacts, frequency-level feature, this representation allows us to capture the trace of up-sampling operations.
% Additionally, we introduce NPRP, an extension of our approach. to
%The overall architecture of NPR and NPRP are shown in Fig. \ref{}.

\subsection{Problem setup}

%Our primary focus centers on the domain of Generalizable Deepfake Dete}ction. 
The overarching objective of Generalizable Deepfake Detection is to develop a universal detector capable of accurately identifying deepfake images, even when faced with limitations in the availability of diverse training sources.

%Our primary focus lies within the realm of Generalizable Deepfake Detection. Our objective is to construct a universal detector capable of accurately identifying deepfake images even when confronted with constrained training sources. 
In the given context, we consider a real-world image scenario denoted as $X$, which is sampled from $n$ different sources:
\begin{equation}
\begin{split}
    &X = \{  X_{1}, X_{2}, \ldots ,X_{i}, \ldots, X_{n} \}, \\ 
    &X_{i} =  \{ x_{j}^{i}, y_{j}\}_{j=1}^{N_{i}},  
 \end{split}
  \label{eq:eq1}
\end{equation}
where $N_i$ represents the number of images originating from the $i$th source $X_{i}$, and $x_{j}^{i}$ is the $j$th image of $X_{i}$. Each image is labeled with $y$, indicating whether it belongs to the category of "real" ($y=0$) or "fake" ($y=1$).

%In the given context, let us consider a real-world image scenario denoted as $X$ sampled from $n$ different sources:
%\begin{equation}
%\begin{split}
%    &X = \{  X_{1}, X_{2}, \ldots ,X_{i}, \ldots, X_{n} \}, \\ 
%    &X_{i} =  \{ x_{j}^{i}, y_{j}\}_{j=1}^{N_{i}},  
% \end{split}
%  \label{eq:eq1}
%\end{equation}
%where $N_i$ represents the number of images originating from the $i$th source $X_{i}$, $x_{j}^{i}$ is the $j$th image of $X_{i}$. Each image is labeled with $y$, indicating whether it belongs to the category of "real"$(y=0)$ or "fake" $(y=1)$. 

Here, we train a binary classifier \(D(\cdot)\), utilizing the training source \(X_{i}\):
\begin{equation}
\begin{split}
 & P_{i} = f(X_{i}), \\
 & D^{i} = \mathop{\arg\min}_{\theta} \ loss(D( P_{i}; \theta),\  y),
\end{split}
\label{eq:eq3}
\end{equation}
where \(f()\) is the representation extractor, \(P_{i}\) is the artifact representation of \(X_{i}\).

%Here we train a binary classifier \(D(\cdot)\), utilizing the training source $X_{i}$:
% \begin{equation}
% \begin{split}
% & P_{i} = f( X_{i} ), \\
% & D^{{i}} = \mathop{\arg\min}_{\theta} \ l(D( P_{i}; \theta),\  y), 
%  \end{split}
%  \label{eq:eq3}
%\end{equation}
%where $f()$ is the representation extractor, $P_{i}$ is artifact representation of $X_{i}$,  $l()$ denote the loss function. 

Our overarching goal is to design a well extractor \(f()\), which extracts a generalized artifact \(P_{i}\) from the training source \(X_{i}\). Subsequently, the generalizable detector \(D^{i}\) can be obtained by training on the artifact \(P_{i}\) originating from \(X_i\), yet it demonstrates robust performance when faced with images from previously unseen sources denoted as \(X_{t}\). The ability to generalize across unseen sources is a crucial objective of our detector representation extractor \(f()\).

 \begin{figure}[ht!]
   \centering
    \includegraphics[scale=0.420]{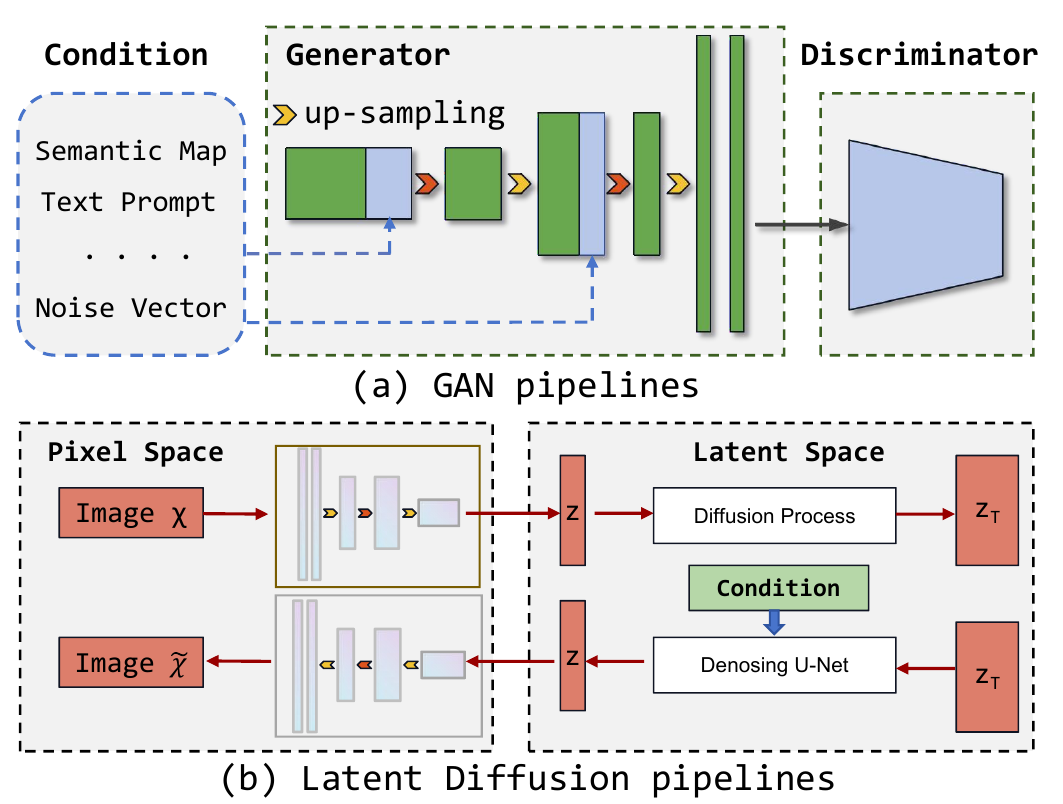}
    \caption{In the pipelines of common generation models, GAN and Diffusion, up-sampling is employed to transform the low-resolution latent space into high resolution. }
    \label{fig:pipelines}
    \vspace{-0.25 cm}
 \end{figure}

 \begin{figure}[ht!]
\vspace{-0.45 cm}
   \centering
    \includegraphics[scale=0.380]{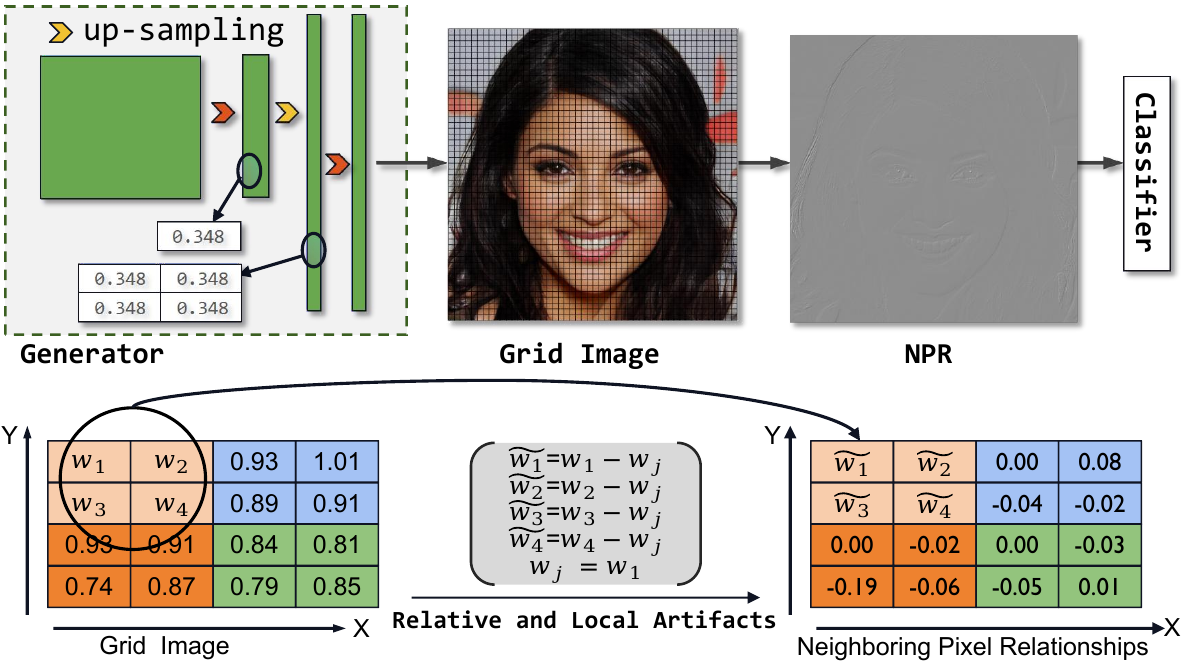}
    \caption{\tft{The overview of Neighboring Pixel Relationships.} We rethink artifacts stemming from the up-sampling component of common generation models. 
The proposed Neighboring Pixel Relationships focus on the local interdependence between image pixels caused by up-sampling operators. The NPR is employed to train detector as artifact representation.}
    \label{fig:pipelines}
  \vspace{-0.25 cm}
 \end{figure}

\subsection{Up-sampling operations in generator pipeline}

Before we dive into the details of the method, let's briefly explore the up-sampling operations commonly used in generator pipelines, such as those in GANs and Diffusions.

\noindent \textbf{GAN pipelines:} 
We present an overview of the fundamental pipeline inherent to Generative Adversarial Networks (GANs), as depicted in Figure \ref{fig:pipelines} (a). 
It comprises two primary constituents, namely the discriminator and the generator. In the context of a GAN, the generator function serves to establish a mapping that originates in a lower-dimensional latent space and extends to the image space.
Within the architecture of the generator, two predominant components are typically incorporated, including convolutional layers and up-sampling layers. 
In these up-sampling layers, their primary function is to accept low-resolution features as input and subsequently generate high-resolution features as their output.
 It is noteworthy to emphasize that while the architectural configurations of GAN models exhibit substantial diversity, the adoption of an upsampling module maintains consistency.
Given the widespread incorporation of up-sampling layers within GAN pipelines, it is pertinent to recognize that the artifacts arising from these up-sampling operations have the potential to enhance the generalization capacity of the detection model.

\noindent \textbf{Diffusion pipelines:} 
Additionally, the Diffusion pipeline is illustrated in Figure \ref{fig:pipelines} (b). 
Diffusion models consist of a forward diffusion process and a reverse diffusion process. The forward process is a Markov chain where noise is gradually added to the data when sequentially sampling the latent variables. In the reverse process, parameterized by another Gaussian transition, a U-Net network is used to denoise the data samples to reconstruct clean images.
Recently, diffusion models include two structures: diffusion with U-Net and latent diffusion models. In diffusion with U-Net, the U-Net model  \cite{ronneberger2015u} is employed to estimate the noise component from a noisy image. During inference time, diffusion models sample noise and gradually reduce the noise level until obtaining a clean image. 
The latent diffusion models include an encoder, denoising U-Net, and decoder. It uses a U-Net to perform diffusion in a latent domain and then decodes the latent signal with a decoder to generate an image. Although the processes of generation in the diffusion model and GAN are different, the decoder of the diffusion model also widely adopts up-sampling layers to generate images.

\subsection{ Neighboring pixel relationships } 
Building upon the above analysis of generation pipelines, we observe that up-sampling operations are commonly employed in current image generation techniques, including GANs and Diffusions. While existing research has delved into studying global up-sampling artifacts in the frequency domain \cite{Frank, Durall}, Jeong \et \cite{jeong2022frepgan}  have discovered that frequency-based artifacts are insufficient for achieving generalization detection, given the diverse patterns in the frequency domain of GANs. 
In this context, we reconsider the up-sampling layer in popular generation models and introduce the concept of local up-sampling artifacts in the spatial domain.

%Based on the above analysis of generation pipeline, we can find out that the upsampling operations commonly used in the current image generation techniques, including of GANs and Diffusions. 
%Existing researches has study the global up-sampling artifact in the frequency domain \cite{Frank, Durall}. However, Jeong \et \cite{jeong2022frepgan} find that  frequency-based artifacts are not up to the generalization detection, since various patterns of frequency domain in GANs. 
%We also will prove that the frequency-level artifacts cannot generalize from GAN to diffusion models in the section . 
%Here we rethink up-sampling layer in popular generation models, and develop the local up-sampling artifact from spatial domain. 
%Two most commonly used up-sampling modules in the generator are traditional upsampling methods, such as nearest neighbor or bilinear upsampling, and deconvolution layer. 

We focus on the portion of the generator near the output images, consisting of an up-sampling layer $up$ with $l$ scale, convolutional layers $conv$ with activate functions, input feature maps $x \in \mathbb{R}^{W \times H \times C}$, and the output images $I \in \mathbb{R}^{(l \times W) \times (l \times H) \times 3}$.
 \begin{equation}
 \begin{split}
 & \hat{x} = up( x ), \\
 & I = conv(\hat{x}), 
  \end{split}
  \label{eq:eq3}
\end{equation}
where $\hat{x} \in \mathbb{R}^{(l \times W) \times (l \times H) \times C}$ is the up-scaled feature map.  
%We then divide the image $I$ and $\hat{x}$ into $\frac{W}{l} \times \frac{H}{l}$ grids. Each gird is the $l \times l$ patches. 
We then divide the image $I$ and $\hat{x}$ into $W \times H$ grids. Each gird is the $l \times l$ patches. 
Let $V_I$ and $V_{\hat{x}}$ denote grids set of $I$ and $\hat{x}$, respectively. The $v_I^{c} \in V_I $ and $v_{\hat{x}}^{c} \in V_{\hat{x}}$ indicate a gird of $I$ and $\hat{x}$, respectively. Most of generators commonly  employ an up-sampling layer with $l=2$ scale.

%Two most commonly used up-sampling modules in the generator are traditional upsampling methods, such as nearest neighbor or bilinear upsampling, and deconvolution layer. 
%%Here we first consider the nearest neighbor upsampling, and then generalize to others. 
%We consider the part of the generator near the output images, consisting of up-sampling layer $up$ with $l$ scale, convolutional layers $conv$ with activate functions, input feature maps $x \in \mathbb{R}^{W \times H \times C}$, and the output images $I \in \mathbb{R}^{(l \times W) \times (l \times H) \times 3}$.
% \begin{equation}
% \begin{split}
% & \hat{x} = up( x ), \\
% & I = conv(\hat{x}), 
%  \end{split}
%  \label{eq:eq3}
%\end{equation}
%where $\hat{x} \in \mathbb{R}^{(l \times W) \times (l \times H) \times C}$ is the up-scaled feature map.  
%We then divide the image $I$ and $\hat{x}$ into $\frac{W}{l} \times \frac{H}{l}$ grids. Each gird is the $l \times l$ patches. 
%Let $V_I$ and $V_{\hat{x}}$ denote grids set of $I$ and $\hat{x}$, respectively. $v_I^{c} \in V_I $ and $v_{\hat{x}}^{c} \in V_{\hat{x}}$ indicate each gird. Most of generators employ up-sampling layer with $l=2$ scale. 

The elements of $v_{\hat{x}}^{c}$ exhibit a strong correlation generated by the up-sampling layer.
For instance, when adopting nearest neighbor interpolation as the up-sampling layer, the elements of  $v_{\hat{x}}^{c}$ share same value. 
 Here are some key characteristics:
1) The elements of $v_{\hat{x}}^{c}$ has strong correlation generated by upsampling layer,
2) The function $conv$ is fixed during inference,
3) The function $conv$ is translation invariance.
Consequently, the correlation of elements is presented in $v_{\hat{x}}^{c}$. We capture the correlation of local pixels in $v_{\hat{x}}^{c}$ as the up-sampling artifacts.  

%The elements of $v_{\hat{x}}^{c}$ has strong correlation generated by upsampling layer.  For example, when adopting nearest neighbor interpolation as the up-sampling layer, the elements of $v_{\hat{x}}^{c}$ share same value. 
%Here are some characteristic:
%1) The elements of $v_{\hat{x}}^{c}$ has strong correlation generated by upsampling layer,
%2) The function $conv$ is fixed during inference,
%3) The function $conv$ is translation invariance.
%Then, the correlation of elements is presented in $v_{\hat{x}}^{c}$. We capture the correlation of local pixels in $v_{\hat{x}}^{c}$ as the up-sampling artifacts.  

Specifically,  the differences in each $v_{\hat{x}}^{c}$ are extracted  as artifacts representation, as following: 
 \begin{equation}
 \begin{split}
 &       v_{I}^{c} = \{w_1, ...,  w_i, ..., w_n \},  n = l{\times}l\\
 &\hat{v}_{I}^{c} = \{w_1-w_j,  ...,  w_i-w_j, ..., w_n-w_j \},  1  \leq  j \leq n, 
  \end{split}
  \label{eq:eq4}
\end{equation}
where $w_i$ is the elements of $v_{I}^{c}$, $\hat{v}_{I}^{c}$ denotes the neighboring pixel relationships of $v_{\hat{x}}^{c}$. 
We adopt subtraction to capture relative relationship of pixels in ${v}_{I}^{c}$. 
The $w_j$ can be employed by any element in $v_{I}^{c}$. 
The NPR of the whole image is  the set of all grids $\hat{v}_{I}^{c}$. Our NPR set $l$ and $j$ to 2 and 1, respectively.  
In the Section \ref{sec:experiments}, We will discuss the effect of $l$ and  $w_j$, and explore the possibility of replacing $w_j$ with max or mean of $v_{I}^{c}$. 

We employ the proposed neighboring pixel relationships $\hat{v}_{I}^{c}$ as the artifacts representation to train the classifier for deepfake detection. 
The NPR captures the local relative correlation between pixels in local patches. This correlation, presented in the image domain, derives from the up-sampling layer and benefits from the translation invariance of the convolutional layer. 
The relative and local nature of the proposed up-sampling artifacts allows the neighboring pixel relationship to be generalized to unknown sources.

%Thanks to \ccn{ the relative and local} nature of the proposed up-sampling artifacts, the neighboring pixel relationships are able to generalize to unseen sources.

%The proposed neighboring pixel relationships capture local relative correlation between pixels in local patches. This correlation presented in image domain derive from the up-sampling layer, and benefit from translation invariance of convolutional layer. 
%Thank to relative and local of the proposed up-sampling artifacts, the  neighboring pixel relationships is able to generalize to unseen sources.

%their primary function is to accept low-resolution features as input and subsequently generate high-resolution images as their output. 
%Therefore, since most pixels in images are correlated to each other, i.e., colors mostly change gradually

%\ccn{add summary}

%\subsection{ Neighboring pixel relationships pyramid }

% The bellowing is copied from 'Leveraging Frequency Analysis for Deep Fake Image Recognition'
%To analyze if this pattern in the spectral domain is a common occurrence for different GAN types and implementations, or simply a fault specific to the StyleGAN instance we studied, we selected four different architectures, namely BigGAN

\section{Experiments}
\label{sec:experiments}

\subsection{Settings}
\label{settings}
\noindent \textbf{Training Dataset:}

To ensure a consistent basis for comparison, we employ the training set of ForenSynths \cite{wang2020cnn} to train the detectors, following baselines \cite{wang2020cnn,jeong2022bihpf,jeong2022frepgan}. 
The training set consists of 20 distinct  categories, each comprising 18,000 synthetic images generated using ProGAN, alongside an equal number of real images sourced from the LSUN dataset. In line with previous research  \cite{jeong2022bihpf,jeong2022frepgan}, we adopt specific 4-class training settings, denoted as $(car, cat, chair, horse)$.

\noindent\textbf{Testing Dataset:}

To assess the generalization ability of the proposed method on the real-world scenarios, we adopt various real images and diverse GAN and Diffusions models. The evaluation dataset consists of \tft{five datasets} containing \tft{28 generation models}.

\noindent\textbf{$\bullet$ 8 models from ForenSynths\cite{wang2020cnn} :} The test set includes fake images generated by 8 generation models~\footnote{ProGAN\cite{karras2018progressive}, StyleGAN\cite{karras2019style}, StyleGAN2\cite{karras2020analyzing}, BigGAN\cite{BigGAN}, CycleGAN \cite{CycleGAN}, StarGAN \cite{choi2018stargan}, GauGAN\cite{GauGAN} and Deepfake \cite{Deepfake}}. Real images are sampled from 6 datasets (LSUN\cite{yu2015lsun}, ImageNet\cite{russakovsky2015imagenet}, CelebA\cite{CelebA}, CelebA-HQ\cite{karras2018progressive}, COCO\cite{coco}, and FaceForensics++\cite{Deepfake}).

% Firstly, we employ the test set of ForenSynths\nocite{wang2020cnn} for evaluation. It includes fake images generated by 8 generation model~\footnote{ProGAN \cite{karras2018progressive}, StyleGAN \cite{karras2019style}, StyleGAN2 \cite{karras2020analyzing}, BigGAN \cite{BigGAN}, CycleGAN \cite{CycleGAN}, StarGAN \cite{choi2018stargan}, GauGAN \cite{GauGAN} and Deepfake \cite{Deepfake}}. The real images are sampled from 6 datasets~\footnote{LSUN \cite{yu2015lsun}, ImageNet \cite{russakovsky2015imagenet}, CelebA \cite{CelebA}, CelebA-HQ \cite{karras2018progressive}, COCO \cite{coco}, and FaceForensics++ \cite{Deepfake}}. 

\noindent\textbf{$\bullet$ 9 GANs from  Self-Synthesis:} To replicate the unpredictability of wild scenes, we extend our evaluation by collecting images generated by 9 additional GANs~\footnote{AttGAN\cite{AttGAN}, BEGAN\cite{began}, CramerGAN\cite{CramerGAN}, InfoMaxGAN\cite{InfoMaxGAN}, MMDGAN\cite{MMDGAN}, RelGAN\cite{RelGAN}, S3GAN\cite{S3GAN}, SNGAN\cite{SNGAN}, and STGAN\cite{STGAN}}. 
There are 4K test images for each model, with equal numbers of real and fake images.

%Additionally, to replicate the unpredictability of wild scenes, we extend our evaluation by collecting images generated by 9 additional GANs~\footnote{AttGAN\cite{AttGAN}, BEGAN\cite{began}, CramerGAN\cite{CramerGAN}, InfoMaxGAN\cite{InfoMaxGAN}, MMDGAN\cite{MMDGAN}, RelGAN\cite{RelGAN}, S3GAN\cite{S3GAN}, SNGAN\cite{SNGAN}, and STGAN\cite{STGAN}}. There are 36K test images, with equal numbers of real and fake images.

\noindent\textbf{$\bullet$ 8 Diffusions from DIRE \cite{Wang_2023_ICCV}: } To expand the testing scope, we adopt the diffusions dataset of DIRE \cite{Wang_2023_ICCV} for evaluation, including ADM \cite{dhariwal2021diffusion}, DDPM \cite{ho2020denoising}, IDDPM \cite{nichol2021improved}, LDM \cite{rombach2022high}, PNDM \cite{liu2022pseudo}, Vqdiffusion \cite{gu2022vector}, Stable Diffusion v1 \cite{rombach2022high}, Stable Diffusion v2 \cite{rombach2022high}. 
The real images are sampled from LSUN \cite{yu2015lsun} and ImageNet\cite{russakovsky2015imagenet} datasets.

% To expand the testing scope, we further adopt diffusions dataset of DIRE \cite{Wang_2023_ICCV} for evaluation, including  ADM \cite{dhariwal2021diffusion}, DDPM \cite{ho2020denoising}, IDDPM \cite{nichol2021improved}, LDM \cite{rombach2022high}, PNDM \cite{liu2022pseudo}, Vqdiffusion \cite{gu2022vector}, Stable Diffusion v1 \cite{rombach2022high}, Stable Diffusion v2 \cite{rombach2022high}.

\noindent\textbf{$\bullet$ 4 Diffusions from Ojha \cite{ojha2023towards}: } This test set contains images generated from ADM  \cite{dhariwal2021diffusion}, Glide \cite{nichol2021glide}, DALL-E-mini \cite{ramesh2021zero}, LDM \cite{rombach2022high}. 
It adopts images of  LAION\cite{schuhmann2021laion} and ImageNet\cite{russakovsky2015imagenet} datasets as the real data.

\noindent\textbf{$\bullet$ 5 Diffusions from Self-Synthesis: } Moreover, we sample test images generated from diffusion models using 1000 diffusion steps, namely DDPM\cite{ho2020denoising}, IDDPM\cite{nichol2021improved}, ADM\cite{dhariwal2021diffusion}, collect images of  Midjourney\footnote{discord.com/channels/662267976984297473}, and DALLE\cite{ramesh2021zero}\footnote{discord.com/channels/974519864045756446} from social platform Discord.

More detailed information on the test set is given in the supplementary material.

%\noindent\textbf{Evaluation Metrics:}

\noindent\textbf{Implementation Details:}
We design a lightweights CNN network using convolutional layer and Resnet\cite{he2016deep} block as the classifiers for NPR  with 1.44 million parameters. The detector is trained using the Adam optimizer\cite{kingma2015adam} with a learning rate of \(2 \times 10^{-4}\), a batch size of 32. %, and 50 epochs. 
Our method is implemented using the PyTorch on Nvidia GeForce RTX 3090 GPU.
To assess the performance of the proposed method, we follow the evaluation metrics used in the baselines\cite{jeong2022bihpf,jeong2022frepgan,ojha2023towards}, which include the average precision score (A.P.) and accuracy (Acc.). 

%We design two lightweights using convolutional layer and Resnet\cite{he2016deep} block as the classifiers for NPR and NPRP with 1.44 M  and \ccn{1.* M} parameters, respectively. The detector is trained using the Adam optimizer\cite{kingma2015adam} with a learning rate of \(2 \times 10^{-4}\) and he batch size 32. 
%Our method is implemented using the PyTorch on Nvidia GeForce RTX 3090 GPU.
%To assess the performance of the proposed method, we follow the evaluation metrics used in the baselines\cite{jeong2022bihpf,jeong2022frepgan,ojha2023towards}, which include the average precision score (A.P.) and accuracy (Acc.). 

%To accommodate different DIOs, we modify the number of channels in the first convolutional layer of ResNet50. 
%For the DIO framework, we set the batch size to 128 and train for 100 epochs. 
%As for the Multi-DIOs (MDIO), the batch size is set to 32, and training is performed for 40 epochs. 
%We apply a learning rate decay strategy, reducing the learning rate by ten percent after every ten epochs for DIO and four epochs for MDIO, respectively. 

\noindent\textbf{Baselines:}
We perform comparisons the proposed NPR with existing deepfake detection works, including   
CNNDetection(CVPR2020) \cite{wang2020cnn},
Frank(PRML 2020) \cite{Frank},
Durall(CVPR 2020) \cite{Durall},
Patchfor(ECCV 2020) \cite{chai2020makes},
F3Net(ECCV 2020) \cite{qian2020thinking},
SelfBland(CVPR 2022)\cite{shiohara2022detecting}, 
GANDetection(ICIP 2022) \cite{mandelli2022detecting},
BiHPF(WACV 2022) \cite{jeong2022bihpf}, 
FrePGAN(AAAI 2022)\cite{jeong2022frepgan},
LGrad(CVPR 2023) \cite{Tan2023CVPR},
Ojha(CVPR 2023) \cite{ojha2023towards}.
We re-implement baselines \cite{wang2020cnn, Frank, Durall, chai2020makes, qian2020thinking, ojha2023towards} with the official codes using 4-classes training setting, and adopt the official pretrained models of baselines\cite{shiohara2022detecting, mandelli2022detecting, Tan2023CVPR}.

\begin{table*}[!ht]
 \vspace{-0.2 cm}
    \centering
\resizebox{\textwidth}{24mm}{
    \begin{tabular}{l  c c c c c c c c c c c c c c c c | c c}
    \bottomrule \hline
       \multirow{2}*{Method} & \multicolumn{2}{c}{ProGAN}& \multicolumn{2}{c}{StyleGAN}& \multicolumn{2}{c}{StyleGAN2}& \multicolumn{2}{c}{BigGAN}& \multicolumn{2}{c}{CycleGAN}& \multicolumn{2}{c}{StarGAN}& \multicolumn{2}{c}{GauGAN}& \multicolumn{2}{c|}{Deepfake}& \multicolumn{2}{c}{Mean}\\

         \cline{2-19} ~   & Acc. & A.P. & Acc. & A.P. & Acc. & A.P. & Acc. & A.P. & Acc. & A.P. & Acc. & A.P. & Acc. & A.P.  & Acc. & A.P.  & Acc. & A.P. \\ \bottomrule \hline
        CNNDetection\cite{wang2020cnn}      & 91.4 & 99.4 & 63.8 & 91.4 & 76.4 & 97.5 & 52.9 & 73.3 & 72.7 & 88.6 & 63.8 & 90.8 & 63.9 & 92.2 & 51.7 & 62.3 & 67.1  &86.9 \\ 
%        High-Freq                              &  Freq  & 4 & 98.9 & 100.0& 74.4 & 98.3 & 68.8 & 97.3 & 75.2 & 92.1 & 71.0 & 87.9 & 92.7 & 100.0& 75.5 & 86.5 & 57.0 & 74.9 & 76.7 & 92.1 \\
        Frank\cite{Frank}                & 90.3 & 85.2 & 74.5 & 72.0 & 73.1 & 71.4 & 88.7 & 86.0 & 75.5 & 71.2 & 99.5 & 99.5 & 69.2 & 77.4 & 60.7 & 49.1 & 78.9 & 76.5 \\ 
        Durall\cite{Durall}              & 81.1 & 74.4 & 54.4 & 52.6 & 66.8 & 62.0 & 60.1 & 56.3 & 69.0 & 64.0 & 98.1 & 98.1 & 61.9 & 57.4 & 50.2 & 50.0 & 67.7  & 64.4\\ 
        Patchfor\cite{chai2020makes}    & 97.8 & 100.0 & 82.6 & 93.1 & 83.6 & 98.5 & 64.7 & 69.5 & 74.5 & 87.2 & 100.0 & 100.0 & 57.2 & 55.4 & 85.0 & 93.2 & 80.7 & 87.1  \\
        F3Net\cite{qian2020thinking}      & 99.4 & 100.0& 92.6 & 99.7 & 88.0 & 99.8 & 65.3 & 69.9 & 76.4 & 84.3 & 100.0& 100.0& 58.1 & 56.7 & 63.5 & 78.8 & 80.4 & 86.2 \\
        SelfBland\cite{shiohara2022detecting}  & 58.8 & 65.2 & 50.1 & 47.7 & 48.6 & 47.4 & 51.1 & 51.9 & 59.2 & 65.3 & 74.5 & 89.2 & 59.2 & 65.5 & 93.8 & 99.3 & 61.9 & 66.4  \\
GANDetection\cite{mandelli2022detecting} & 82.7 & 95.1 & 74.4 & 92.9 & 69.9 & 87.9 & 76.3 & 89.9 & 85.2 & 95.5 & 68.8 & 99.7 & 61.4 & 75.8 & 60.0 & 83.9 & 72.3 & 90.1  \\
%        Resnet50                               &  Image & 4 & 99.8 & 100.0& 82.6 & 90.3 & 74.9 & 98.4 & 59.8 & 63.4 & 63.5 & 71.1 & 100.0& 100.0& 55.6 & 55.2 & 73.4 & 95.7 & 76.2 & 84.3 \\
        BiHPF\cite{jeong2022bihpf}       & 90.7 & 86.2 & 76.9 & 75.1 & 76.2 & 74.7 & 84.9 & 81.7 & 81.9 & 78.9 & 94.4 & 94.4 & 69.5 & 78.1 & 54.4 & 54.6 & 78.6  & 77.9\\
        FrePGAN\cite{jeong2022frepgan}    & 99.0 & 99.9 & 80.7 & 89.6 & 84.1 & 98.6 & 69.2 & 71.1 & 71.1 & 74.4 & 99.9 & 100.0& 60.3 & 71.7 & 70.9 & 91.9 & 79.4  & 87.2\\ 
        LGrad \cite{Tan2023CVPR}         & 99.9 & 100.0& 94.8 & 99.9 & 96.0 & 99.9 & 82.9 & 90.7 & 85.3 & 94.0 & 99.6 & 100.0& 72.4 & 79.3 & 58.0 & 67.9 & 86.1 & 91.5 \\
        Ojha \cite{ojha2023towards}    & 99.7 & 100.0& 89.0 & 98.7 & 83.9 & 98.4 & 90.5 & 99.1 & 87.9 & 99.8 & 91.4 & 100.0& 89.9 & 100.0& 80.2 & 90.2 & {\cblue{89.1}} & {\cred{98.3}} \\
       % FreqNet                                &  Freq  & 4 & 99.6 & 100.0& 90.2 & 99.7 & 88.0 & 99.5 & 90.5 & 96.0 & 95.8 & 99.6 & 85.7 & 99.8 & 93.4 & 98.6 & 88.9 & 94.4 & \tft{91.5} & \tft{98.5}\\
 \rowcolor{light-gray} NPR(our)                & 99.8 & 100.0& 96.3 & 99.8 & 97.3 & 100.0& 87.5 & 94.5 & 95.0 & 99.5 & 99.7 & 100.0& 86.6 & 88.8 & 77.4 & 86.2 & \cred{92.5} & \cblue{96.1}\\
% \rowcolor{light-gray} NPRP                    & ---- & ---- & ---- & ---- & ---- & ---- & ---- & ---- & ---- & ---- & ---- & ---- & ---- & ---- & ---- & ---- & ---- & ----  \\\underline

\bottomrule
    \end{tabular}
}
  \caption{\tft{Cross-GAN-Sources Evaluation on the test set of ForenSynths\cite{wang2020cnn}.} The results of \cite{wang2020cnn,Frank,Durall,jeong2022bihpf,jeong2022frepgan} are from \cite{jeong2022frepgan, jeong2022bihpf}. 
  {\cred{Red}} and {\cblue{Blue}} represent the best and second-best performance, respectively. }% Our methods perform favorably among the works compared.
  \label{tab:SOTA1}
     \vspace{-0.25 cm}
\end{table*}

\begin{table*}[!ht]
 \vspace{-0.2 cm}
    \centering
\resizebox{\textwidth}{19mm}{
    \begin{tabular}{l  c c c c c c c c c c c c c c c c c c| c c}
    \bottomrule \hline
      %\multirow{3}*{Method} &\multicolumn{18}{c}{ Test Models}\\ 
       \multirow{2}*{Method} & \multicolumn{2}{c}{AttGAN}& \multicolumn{2}{c}{BEGAN}& \multicolumn{2}{c}{CramerGAN}& \multicolumn{2}{c}{InfoMaxGAN}& \multicolumn{2}{c}{MMDGAN}& \multicolumn{2}{c}{RelGAN}& \multicolumn{2}{c}{S3GAN}& \multicolumn{2}{c}{SNGAN}&  \multicolumn{2}{c|}{STGAN}& \multicolumn{2}{c}{Mean}\\
         \cline{2-21} ~   & Acc. & A.P. & Acc. & A.P. & Acc. & A.P. & Acc. & A.P. & Acc. & A.P. & Acc. & A.P. & Acc. & A.P. & Acc. & A.P. & Acc. & A.P. & Acc. & A.P.\\ \bottomrule \hline
CNNDetection\cite{wang2020cnn}                 & 51.1 & 83.7 & 50.2 & 44.9 & 81.5 & 97.5 & 71.1 & 94.7  & 72.9 & 94.4 & 53.3 & 82.1 & 55.2 & 66.1 & 62.7 & 90.4 & 63.0 & 92.7 & 62.3 & 82.9 \\
Frank\cite{Frank}                                            & 65.0 & 74.4 & 39.4 & 39.9 & 31.0 & 36.0 & 41.1 & 41.0 & 38.4 & 40.5 & 69.2 & 96.2 & 69.7 & 81.9 & 48.4 & 47.9 & 25.4 & 34.0 & 47.5 & 54.7 \\
Durall\cite{Durall}                                            & 39.9 & 38.2 & 48.2 & 30.9 & 60.9 & 67.2 & 50.1 & 51.7 & 59.5 & 65.5 & 80.0 & 88.2 & 87.3 & 97.0 & 54.8 & 58.9 & 62.1 & 72.5 & 60.3 & 63.3 \\
Patchfor\cite{chai2020makes}                        & 68.0 & 92.9 & 97.1 & 100.0 & 97.8 & 99.9 & 93.6 & 98.2 & 97.9 & 100.0 & 99.6 & 100.0 & 66.8 & 68.1 & 97.6 & 99.8 & 92.7 & 99.8 & \cblue{90.1} & 95.4 \\
F3Net\cite{qian2020thinking}   & 85.2 & 94.8 & 87.1 & 97.5 & 89.5 & 99.8 & 67.1 & 83.1 & 73.7 & 99.6 & 98.8 & 100.0 & 65.4 & 70.0 & 51.6 & 93.6 & 60.3 & 99.9 & 75.4 & 93.1 \\
SelfBland\cite{shiohara2022detecting}           & 63.1 & 66.1 & 56.4 & 59.0 & 75.1 & 82.4 & 79.0 & 82.5 & 68.6 & 74.0 & 73.6 & 77.8 & 53.2 & 53.9 & 61.6 & 65.0 & 61.2 & 66.7 & 65.8 & 69.7 \\
GANDetection\cite{mandelli2022detecting}    & 57.4 & 75.1 & 67.9 & 100.0 & 67.8 & 99.7 & 67.6 & 92.4 & 67.7 & 99.3 & 60.9 & 86.2 & 69.6 & 83.5 & 66.7 & 90.6 & 69.6 & 97.2 & 66.1 & 91.6 \\
LGrad \cite{Tan2023CVPR}   & 68.6 & 93.8 & 69.9 & 89.2 & 50.3 & 54.0 & 71.1 & 82.0 & 57.5 & 67.3 & 89.1 & 99.1 & 78.5 & 86.0 & 78.0 & 87.4 & 54.8 & 68.0 & 68.6 & 80.8\\
%DIO-VGG & 87.5 & 95.4 & 79.1 & 97.7 & 95.2 & 98.0 & 92.2 & 95.1  & 95.0 & 97.2 & 96.4 & 99.8 & 70.9 & 72.7 & 92.8 & 95.9 & 78.8 & 98.1 & 87.5 & 94.4 \\
Ojha \cite{ojha2023towards}  & 78.5 & 98.3 & 72.0 & 98.9 & 77.6 & 99.8 & 77.6 & 98.9 & 77.6 & 99.7 & 78.2 & 98.7 & 85.2 & 98.1 & 77.6 & 98.7 & 74.2 & 97.8 & 77.6 & \cred{98.8}\\
% \rowcolor{light-gray}NPR(our)    & 83.0 & 96.2 & 99.0 & 99.8 & 98.7 & 99.0 & 51.8 & 70.7 & 94.5 & 98.3 & 98.6 & 99.0 & 99.6 & 100.0 & 79.0 & 80.0 & 88.8 & 97.4 & \cred{98.0} & \cred{100.0}  \\
 \rowcolor{light-gray}NPR(our)    &  83.0 & 96.2 & 99.0 & 99.8 & 98.7 & 99.0 & 94.5 & 98.3 & 98.6 & 99.0 & 99.6 & 100.0 & 79.0 & 80.0 & 88.8 & 97.4 & 98.0 & 100.0 & \cred{93.2} & \cblue{96.6} \\
% \rowcolor{light-gray}NPRP(our)    & - & - & - & - & - & - & - & - & - & - & - & - & - & - & - & - & - & - & - & -  \\

\bottomrule
    \end{tabular}
}
  \caption{\textbf{Cross-GAN-Sources Evaluation on the Self-Synthesis 9 GANs dataset.}}
%  \tft{Bold} and \underline{underline} represent the best and second-best performance, respectively.}
  \label{tab:SOTA2}
          \vspace{-0.25 cm}
\end{table*}

\begin{table*}[!ht]
\vspace{-0.2 cm}
    \centering
\resizebox{\textwidth}{23mm}{
    \begin{tabular}{l  c c c c c c c c c c c c c c c c | c c}
    \bottomrule \hline
       \multirow{2}*{Method} & \multicolumn{2}{c}{ADM}& \multicolumn{2}{c}{DDPM}& \multicolumn{2}{c}{IDDPM}& \multicolumn{2}{c}{LDM}& \multicolumn{2}{c}{PNDM}& \multicolumn{2}{c}{VQ-Diffusion} & \multicolumn{2}{c}{\makecell[c]{\small{Stable} \\ \small{Diffusion v1} } }   & \multicolumn{2}{c}{\makecell[c]{\small{Stable} \\ \small{Diffusion v2} } }    & \multicolumn{2}{c}{Mean}\\
         \cline{2-19} ~   & Acc. & A.P. & Acc. & A.P. & Acc. & A.P. & Acc. & A.P. & Acc. & A.P. & Acc. & A.P. & Acc. & A.P.  & Acc. & A.P.  & Acc. & A.P. \\ \bottomrule \hline
%       Resnet50                          & ---- & ---- & ---- & ---- & ---- & ---- & ---- & ---- & ---- & ---- & ---- & ---- & ---- & ---- \\
        CNNDetection\cite{wang2020cnn}                            & 53.9 & 71.8 & 62.7 & 76.6 & 50.2 & 82.7 & 50.4 & 78.7 & 50.8 & 90.3 & 50.0 & 71.0 & 38.0 & 76.7 & 52.0 & 90.3 & 51.0 & 79.8 \\
        Frank\cite{Frank}                                                       & 58.9 & 65.9 & 37.0 & 27.6 & 51.4 & 65.0 & 51.7 & 48.5 & 44.0 & 38.2 & 51.7 & 66.7 & 32.8 & 52.3 & 40.8 & 37.5 & 46.0 & 50.2 \\
        Durall\cite{Durall}                                                       & 39.8 & 42.1 & 52.9 & 49.8 & 55.3 & 56.7 & 43.1 & 39.9 & 44.5 & 47.3 & 38.6 & 38.3 & 39.5 & 56.3 & 62.1 & 55.8 & 47.0 & 48.3 \\
        Patchfor\cite{chai2020makes}                                   & 77.5 & 93.9 & 62.3 & 97.1 & 50.0 & 91.6 & 99.5 & 100.0 & 50.2 & 99.9 & 100.0 & 100.0 & 90.7 & 99.8 & 94.8 & 100.0 & 78.1 & 97.8\\
        F3Net\cite{qian2020thinking}                                     & 80.9 & 96.9 & 84.7 & 99.4 & 74.7 & 98.9 & 100.0 & 100.0 & 72.8 & 99.5 & 100.0 & 100.0 & 73.4 & 97.2 & 99.8 & 100.0 & 85.8 & \cblue{99.0} \\
        SelfBland\cite{shiohara2022detecting}                      & 57.0 & 59.0 & 61.9 & 49.6 & 63.2 & 66.9 & 83.3 & 92.2 & 48.2 & 48.2 & 77.2 & 82.7 & 46.2 & 68.0 & 71.2 & 73.9 & 63.5 & 67.6 \\
        GANDetection\cite{mandelli2022detecting}              & 51.1 & 53.1 & 62.3 & 46.4 & 50.2 & 63.0 & 51.6 & 48.1 & 50.6 & 79.0 & 51.1 & 51.2 & 39.8 & 65.6 & 50.1 & 36.9 & 50.8 & 55.4 \\
        LGrad \cite{Tan2023CVPR}                                      & 86.4 & 97.5 & 99.9 & 100.0 & 66.1 & 92.8 & 99.7 & 100.0 & 69.5 & 98.5 & 96.2 & 100.0 & 90.4 & 99.4 & 97.1 & 100.0 & \cblue{88.2} & {98.5} \\
        Ojha \cite{ojha2023towards}                                     & 78.4 & 92.1 & 72.9 & 78.8 & 75.0 & 92.8 & 82.2 & 97.1 & 75.3 & 92.5 & 83.5 & 97.7 & 56.4 & 90.4 & 71.5 & 92.4 & 74.4 & 91.7\\
 \rowcolor{light-gray}NPR (our)                                                                  & 88.6 & 98.9 & 99.8 & 100.0 & 91.8 & 99.8 & 100.0 & 100.0 & 91.2 & 100.0 & 100.0 & 100.0 & 97.4 & 99.8 & 93.8 & 100.0 & \cred{95.3} & \cred{99.8} \\
% \rowcolor{light-gray}NPRP (our)                                                                & - & - & - & -& - & - & -& -& - & -& -& -& - & - & - & - & -  & -  \\

\bottomrule
 \end{tabular}}
  \caption{\textbf{Cross-Diffusion-Sources Evaluation on the test of DiffusionForensics \cite{Wang_2023_ICCV}.}}
  \label{tab:iccvdataset}
        \vspace{-0.25 cm}
\end{table*}

\begin{table*}[!ht]
 \vspace{-0.2 cm}
    \centering
\resizebox{\textwidth}{21mm}{
    \begin{tabular}{l  c c c c c c c c c c c c c c c c | c c}
    \bottomrule \hline
       \multirow{2}*{Method} & \multicolumn{2}{c}{DALLE}& \multicolumn{2}{c}{Glide\_100\_10}& \multicolumn{2}{c}{Glide\_100\_27}& \multicolumn{2}{c}{Glide\_50\_27} & \multicolumn{2}{c}{ADM} & \multicolumn{2}{c}{LDM\_100} & \multicolumn{2}{c}{LDM\_200} & \multicolumn{2}{c}{LDM\_200\_cfg}& \multicolumn{2}{c}{Mean}\\
         \cline{2-19} ~   & Acc. & A.P. & Acc. & A.P. & Acc. & A.P. & Acc. & A.P. & Acc. & A.P. & Acc. & A.P. & Acc. & A.P. & Acc. & A.P. & Acc. & A.P.\\ \bottomrule \hline
%       Resnet50                          & ---- & ---- & ---- & ---- & ---- & ---- & ---- & ---- & ---- & ---- & ---- & ---- & ---- & ---- \\

CNNDetection\cite{wang2020cnn}            & 51.8 & 61.3 & 53.3 & 72.9 & 53.0 & 71.3 & 54.2 & 76.0 & 54.9 & 66.6 & 51.9 & 63.7 & 52.0 & 64.5 & 51.6 & 63.1 & 52.8 & 67.4 \\
Frank\cite{Frank}                         & 57.0 & 62.5 & 53.6 & 44.3 & 50.4 & 40.8 & 52.0 & 42.3 & 53.4 & 52.5 & 56.6 & 51.3 & 56.4 & 50.9 & 56.5 & 52.1 & 54.5 & 49.6 \\
Durall\cite{Durall}                        & 55.9 & 58.0 & 54.9 & 52.3 & 48.9 & 46.9 & 51.7 & 49.9 & 40.6 & 42.3 & 62.0 & 62.6 & 61.7 & 61.7 & 58.4 & 58.5 & 54.3 & 54.0 \\
Patchfor\cite{chai2020makes}                             & 79.8 & 99.1 & 87.3 & 99.7 & 82.8 & 99.1 & 84.9 & 98.8 & 74.2 & 81.4 & 95.8 & 99.8 & 95.6 & 99.9 & 94.0 & 99.8 & 86.8 & 97.2 \\
F3Net\cite{qian2020thinking}      & 71.6 & 79.9 & 88.3 & 95.4 & 87.0 & 94.5 & 88.5 & 95.4 & 69.2 & 70.8 & 74.1 & 84.0 & 73.4 & 83.3 & 80.7 & 89.1 & 79.1 & 86.5 \\
SelfBland\cite{shiohara2022detecting}                          & 52.4 & 51.6 & 58.8 & 63.2 & 59.4 & 64.1 & 64.2 & 68.3 & 58.3 & 63.4 & 53.0 & 54.0 & 52.6 & 51.9 & 51.9 & 52.6 & 56.3 & 58.7 \\
GANDetection\cite{mandelli2022detecting}                   & 67.2 & 83.0 & 51.2 & 52.6 & 51.1 & 51.9 & 51.7 & 53.5 & 49.6 & 49.0 & 54.7 & 65.8 & 54.9 & 65.9 & 53.8 & 58.9 & 54.3 & 60.1 \\
LGrad \cite{Tan2023CVPR}        & 88.5 & 97.3 & 89.4 & 94.9 & 87.4 & 93.2 & 90.7 & 95.1 & 86.6 & 100.0& 94.8 & 99.2 & 94.2 & 99.1 & 95.9 & 99.2 & \cblue{90.9} & \cblue{97.2} \\
Ojha \cite{ojha2023towards}       & 89.5 & 96.8 & 90.1 & 97.0 & 90.7 & 97.2 & 91.1 & 97.4 & 75.7 & 85.1 & 90.5 & 97.0 & 90.2 & 97.1 & 77.3 & 88.6 & 86.9 & 94.5 \\
 \rowcolor{light-gray}NPR (our)                                    & 94.5 & 99.5 & 98.2 & 99.8 & 97.8 & 99.7 & 98.2 & 99.8 & 75.8 & 81.0 & 99.3 & 99.9 & 99.1 & 99.9 & 99.0 & 99.8 & \cred{95.2} & \cred{97.4} \\
% \rowcolor{light-gray}NPRP (our)                       &  ---- & ---- & ---- & ---- & ---- & ---- & ---- & ---- & ---- & ---- & ---- & ---- & ---- & ---- & ---- & ---- & ---- & ----  \\

\bottomrule
    \end{tabular}
    }
  \caption{\textbf{Cross-Diffusion-Sources Evaluation on the diffusion test set of Ojha \cite{ojha2023towards} .}}
  \label{tab:cvprdataset}
        \vspace{-0.25 cm}
\end{table*}

\begin{table*}%[!ht]
 \vspace{-0.2 cm}
 %   \centering

\begin{floatrow}
\capbtabbox{
\resizebox{0.75\textwidth}{22mm}{
    \begin{tabular}{l c c c c c c c c c c | c c}
    \bottomrule \hline
       \multirow{2}*{Method} & \multicolumn{2}{c}{DDPM}& \multicolumn{2}{c}{IDDPM}& \multicolumn{2}{c}{ADM}& \multicolumn{2}{c}{Midjourney} & \multicolumn{2}{c}{DALLE} & \multicolumn{2}{c}{Mean}\\
         \cline{2-13} ~   & Acc. & A.P. & Acc. & A.P. & Acc. & A.P. & Acc. & A.P. & Acc. & A.P. & Acc. & A.P. \\ \bottomrule \hline
CNNDetection\cite{wang2020cnn}                & 50.0 & 63.3 & 48.3 & 52.68 & 53.4 & 64.4 & 48.6 & 38.5 & 49.3 & 44.7 & 49.9 & 52.7 \\
Frank\cite{Frank}                                          & 47.6 & 43.1 & 70.5 & 85.7 & 67.3 & 72.2 & 39.7 & 40.8 & 68.7 & 65.2 & 58.8 & 61.4 \\
Durall\cite{Durall}                                          & 54.1 & 53.6 & 63.2 & 71.7 & 39.1 & 40.8 & 45.7 & 47.2 & 53.9 & 52.2 & 51.2 & 53.1 \\
Patchfor\cite{chai2020makes}                      & 54.1 & 66.3 & 35.8 & 34.2 & 68.6 & 73.7 & 66.3 & 68.8 & 60.8 & 65.1 & 57.1 & 61.6 \\
F3Net\cite{qian2020thinking}                        & 59.4 & 71.9 & 42.2 & 44.7 & 73.4 & 80.3 & 73.2 & 80.4 & 79.6 & 87.3 & 65.5 & 72.9 \\
SelfBland\cite{shiohara2022detecting}                & 55.3 & 57.7 & 63.5 & 62.5 & 57.1 & 60.1 & 54.3 & 56.4 & 48.8 & 47.4 & 55.8 & 56.8\\
GANDetection\cite{mandelli2022detecting}             & 47.3 & 45.5 & 47.9 & 57.0 & 51.0 & 56.1 & 50.0 & 44.7 & 49.8 & 49.7 & 49.2 & 50.6 \\
LGrad \cite{Tan2023CVPR}                          & 59.8 & 88.5 & 45.2 & 46.9 & 72.7 & 79.3 & 68.3 & 76.0 & 75.1 & 80.9 & 64.2 & \cblue{74.3} \\
Ojha \cite{ojha2023towards}                         & 69.5 & 80.0 & 64.9 & 74.2 & 81.3 & 90.8 & 50.0 & 49.8 & 66.3 & 74.6 & \cblue{66.4} & 73.9 \\
 \rowcolor{light-gray}NPR (our)                     & 88.5 & 95.1 & 77.9 & 84.8 & 75.8 & 79.3 & 77.4 & 81.9 & 80.7 & 83.0 & \cred{80.1} & \cred{84.8} \\
% \rowcolor{light-gray}NPRP (our)                   & - & - & - & - & - & - & - & - & - & - & - & -\\
\bottomrule
    \end{tabular}
}}{
      \vspace{-0.3 cm}

  \caption{\textbf{Cross-Diffusion-Sources Evaluation on the Self-Synthesis Diffusion dataset.}
  The images of DALLE and Midjourney are collected from the official channel of Discord. The images of other diffusion models are sampled from official pre-trained models with 1000 diffusion steps.  
        \vspace{-0.3 cm}
  }
  \label{tab:mydiffusion}}
\hspace{-0.8cm}
\capbtabbox{
\resizebox{0.23\textwidth}{20mm}{
\begin{tabular}{l c}
\bottomrule \hline
Method & \makecell[c]{\small{Mean Acc. of } \\ \small{38 sub-testsets} }  \\ 
\bottomrule \hline
CNNDetection\cite{wang2020cnn}           & 57.3 \\
Frank\cite{Frank}                        & 56.8 \\
Durall\cite{Durall}                      & 56.6 \\
Patchfor\cite{chai2020makes}             & \cblue{80.6} \\
F3Net\cite{qian2020thinking}             & 78.1 \\
SelfBland\cite{shiohara2022detecting}    & 61.2 \\
GANDetection\cite{mandelli2022detecting} & 59.5 \\
LGrad \cite{Tan2023CVPR}                 & 80.5 \\
Ojha \cite{ojha2023towards}              & 79.8 \\
 \rowcolor{light-gray}NPR (our)          & \cred{92.2} \\

\bottomrule
    \end{tabular}
}}{ \caption{The mean accuracy of all 28 generation models on five datasets.}
 \label{tab:all}
       \vspace{-0.3 cm}
% \small
}
\end{floatrow}

\end{table*}

\begin{table*}[!ht]
 %\vspace{-0.3 cm}
    \centering
\resizebox{\textwidth}{24.3mm}{
    \begin{tabular}{c c c c c c c c c c c c c c c c c c| c c}
    \bottomrule   \hline
     \multirow{2}*{Size $l\times l$}  &     \multirow{2}*{$w_j$}  & \multicolumn{2}{c}{ProGAN}& \multicolumn{2}{c}{StyleGAN}& \multicolumn{2}{c}{StyleGAN2}& \multicolumn{2}{c}{BigGAN}& \multicolumn{2}{c}{CycleGAN}& \multicolumn{2}{c}{StarGAN}&        \multicolumn{2}{c}{GauGAN}& \multicolumn{2}{c|}{Deepfake}& \multicolumn{2}{c}{Mean}\\ 
      \cline{3-20} ~   &~   & Acc. & A.P. & Acc. & A.P. & Acc. & A.P. & Acc. & A.P. & Acc. & A.P. & Acc. & A.P. & Acc. & A.P. & Acc. & A.P. & Acc. & A.P.\\ \bottomrule \hline

%\hline
%$2\times2$ &  1   & 99.9 & 100.0 & 98.9 & 100.0 & 99.2 & 100.0 & 87.7 & 92.8 & 91.3 & 98.9 & 99.7 & 100.0 & 84.9 & 87.3 & 73.1 & 84.7 & 91.8 & 95.5 \\
$2\times2$ &  $w_1$   & 99.8 & 100.0& 96.3 & 99.8 & 97.3 & 100.0& 87.5 & 94.5 & 95.0 & 99.5 & 99.7 & 100.0& 86.6 & 88.8 & 77.4 & 86.2 & 92.5 & 96.1 \\
$2\times2$ &  $w_2$   & 99.1 & 100.0 & 94.1 & 98.5 & 90.5 & 98.9 & 76.7 & 83.9 & 91.7 & 99.1 & 98.0 & 100.0 & 75.9 & 78.2 & 73.7 & 83.4 & 87.5 & 92.7 \\
$2\times2$ &  $w_3$   & 99.9 & 100.0 & 97.1 & 99.9 & 95.6 & 99.9 & 84.2 & 91.3 & 89.3 & 98.6 & 99.2 & 100.0 & 89.7 & 92.5 & 72.8 & 79.8 & 91.0 & 95.3 \\
$2\times2$ &  $w_4$   & 99.9 & 100.0 & 96.4 & 99.0 & 98.9 & 100.0 & 85.9 & 93.3 & 89.3 & 99.0 & 99.5 & 100.0 & 88.6 & 93.6 & 72.6 & 81.3 & 91.4 & 95.8 \\
$2\times2$ &  $avg(v_{I}^{c})$ & 99.9 & 100.0 & 95.3 & 98.8 & 98.9 & 100.0 & 80.1 & 86.8 & 89.9 & 95.1 & 100.0 & 100.0 & 73.6 & 72.0 & 68.8 & 77.0 & 88.3 & 91.2 \\
$2\times2$ &  $max(v_{I}^{c})$ & 99.9 & 100.0 & 98.8 & 100.0 & 94.9 & 99.9 & 84.2 & 91.8 & 93.1 & 95.8 & 91.3 & 98.6 & 80.0 & 86.0 & 84.9 & 93.0 & 90.9 & 95.6 \\  \hline
$3\times3$ &  $w_1$   & 99.9 & 100.00& 95.8 & 99.9 & 97.8 & 100.0& 82.5 & 88.1 & 81.7 & 93.7 & 96.7 & 99.6 & 81.4 & 87.9 & 79.8 & 83.2 & 89.4 & 94.1 \\
$3\times3$ &  $w_2$   & 99.9 & 100.0 & 92.6 & 99.3 & 95.0 & 99.8 & 83.1 & 92.3 & 86.0 & 97.5 & 99.5 & 100.0& 83.3 & 88.9 & 81.7 & 90.6 & 90.1 & 96.1 \\
$3\times3$ &  $w_3$   & 99.6 & 100.0 & 90.8 & 98.5 & 96.1 & 99.8 & 75.2 & 84.0 & 88.0 & 94.2 & 100.0 & 100.0 & 78.7 & 82.5 & 84.0 & 93.0 & 89.0 & 94.0 \\
$3\times3$ &  $w_4$   & 99.6 & 100.0 & 94.7 & 99.8 & 95.3 & 99.9 & 82.7 & 89.7 & 81.9 & 96.7 & 95.4 & 99.8 & 78.9 & 81.7 & 88.3 & 94.5 & 89.6 & 95.3 \\
$3\times3$ &  $avg(v_{I}^{c})$ & 99.9 & 100.0 & 97.6 & 99.9 & 95.3 & 99.8 & 81.8 & 90.6 & 86.9 & 96.6 & 98.5 & 99.9 & 79.9 & 89.0 & 57.3 & 92.4 & 87.1 & 96.0 \\
$3\times3$ &  $max(v_{I}^{c})$ & 99.9 & 100.0 & 99.3 & 100.0 & 99.3 & 100.0 & 76.3 & 85.2 & 84.5 & 94.1 & 99.0 & 100.0 & 77.8 & 84.8 & 87.9 & 94.7 & 90.5 & 94.8 \\

 \bottomrule 
    \end{tabular}
}
  \caption{
  \textbf{Effect of the hyperparameters of Neighboring Pixel Relationships.}  }
  \label{tab:tab6}
        \vspace{-0.15 cm}

\end{table*}

\subsection{Generalization capability evaluation}

In this section, we demonstrate that the local artifacts representation, Neighboring Pixel Relationships, induced by the up-sampling operations in common generation pipelines, can be easily employed for identifying generated image data. Even a detector trained on a GAN model exhibits the ability to generalize to recently generated diffusion images.

%In this section, we show the local artifacts representation, Neighboring Pixel Relationships, caused by the up-sampling operations in state of the common generation pipelines can be adopt easily to identify the generated images data. Even the detector trained on a GAN model is able to generalize to the recent diffusion-generated images data. 

To analyze if the proposed local up-sampling artifacts is a common occurrence for different generation models, 
we perform the evaluation on a cross-sources dataset comprising images from 28 distinct generation models. 
The details of test set are given in the Section \ref{settings} and the supplementary material. 
The detectors of NPR are trained by the images from ProGAN and subsequently evaluated on 16 GANs, 1 Deepfake, and 11 Diffusion models. 
\underline{We adopt specific 4-} \underline{classes training settings for all experiments in this paper,} \underline{denoted as $ProGAN\mbox{-}(car, cat, chair, horse)$}.

%
%To analyze if the proposed local  up-sampling artifacts is a common occurrence for different  generation models, 
%%To show the generalization ability of the proposed  the local up-sampling artifacts on unseen sources, 
%we perform the evaluation on a cross-sources dataset consisting of images from 26 generation models. The details of test set are given in the Section \ref{settings} and the supplementary material. 
%%In  order to evaluate the generalization ability of the proposed Neighboring Pixel Relationships, 
%The detectors of NPR are trained by the images from ProGAN, and evaluated on the 16 GANs, 1 Deepfake, and 9 Diffusion models. 
%Following the baselines \cite{jeong2022bihpf, jeong2022frepgan, Tan2023CVPR},  we adopt specific 2-classes, and 4-classes training settings, denoted as (chair, horse), (car, cat, chair, horse), respectively. 
%\noindent\textbf{Cross-dataset evaluation.}
%
%\noindent\textbf{Cross-model evaluation.}

%\noindent\textbf{GAN-Sources Evaluation.}
\subsubsection{GAN-Sources Evaluation}

In order to valid the generalization ability on images of GAN sources, two test sets, ForenSynths\cite{wang2020cnn} and self-synthesis GAN datasets, are employed for evaluation. 
These datasets encompass 17 distinct generation models used to test the detection performance of the NPR detector trained on ProGAN images. 
The results are presented in Table \ref{tab:SOTA1} and Table \ref{tab:SOTA2}.

%In order to valid the generalization ability on images of GAN sources, two test sets, ForenSynths\cite{wang2020cnn} and  self-synthesis GAN dataset, are adopted to perform evaluation. 
%In those datasets, 17 generation models are employed to test the detection performance  of NPR detector trained on ProGAN images.  
%The results are reported in Table \ref{tab:SOTA1} and Table \ref{tab:SOTA2}. 

Table \ref{tab:SOTA1} provides a comprehensive overview of the performance of detectors on the test set of ForenSynths\cite{wang2020cnn}. 
The Neighboring Pixel Relationships (NPR) outperforms its counterparts, showcasing higher mean accuracy (Acc.) and comparable mean average precision (A.P.) metrics. 
Particularly noteworthy are the mean accuracy values of NPR, which reach 92.5\%.
It is worth emphasizing the remarkable superiority of NPR over the current state-of-the-art methods, LGrad and Ojha. In terms of mean accuracy, NPR surpasses LGrad and Ojha by 6.4\% and 3.4\%, respectively, underscoring its efficacy in generalizable deepfake detection.

%In Table \ref{tab:SOTA1}, the detectors are trained with 2-classes, and 4-classes settings, respectively. 
%The NPR surpasses its counterparts in terms of mean Acc. and achieves comparable mean A.P. metric. 
%Notably, the mean Acc. values of NPR is 90.2\% and 92.5\%, respectively. 
%Our NPR outperforms the current state-of-the-art LGrad and Ojha by 6.4\% and 3.4\% in terms of mean Acc. metric.

To further assess the generalization ability of Neighboring Pixel Relationships (NPR) across GAN-sources, we expanded the evaluation to include results from 9 additional GAN models, as presented in Table \ref{tab:SOTA2}. 
The results demonstrate the consistent outperformance of NPR in terms of generalization performance on GAN-sources. 
%NPR achieves an impressive average accuracy of 93.2\%, substantially surpassing the best-performing baselines, Ojha \cite{ojha2023towards} and LGrad \cite{Tan2023CVPR}, which attain accuracy values of 77.6\% and 68.6\%, respectively.
NPR achieves an impressive average accuracy of 93.2\%, substantially outperforming the best-performing baselines, Patchfor \cite{chai2020makes} and Ojha \cite{ojha2023towards}, which attain accuracy values of 90.1\% and 77.6\%, respectively.

%In addition, we collect more GANs to evaluate the generalization ability of NPR on the GAN-sources. The results of 9 GANs are shown in Table \ref{tab:SOTA2}. The 4-classes settings is adopted as training setting.
%It can be observed that our NPR still obtain outperformance generalization performance on the GAN-sources. 
%The NPR gets an average accuracy of 98.0\%, while the best performing baselines, Ojha \cite{ojha2023towards}, LGrad \cite{Tan2023CVPR}, achieve 77.6\% and 75.4\%, respectively.
%%We evaluate the effectiveness of our NPR on the GAN-sources set, as shown in Table \ref{tab:SOTA1} and Table \ref{tab:SOTA2}. The detectors trained with 2-classes, and 4-classes settings are employed to perform GAN-sources evaluation.  

The results obtained across 17 diverse generation models underscore the remarkable generalization capability of the proposed artifacts representation derived from up-sampling operations. Notably, training the detector on ProGAN images enables Neighboring Pixel Relationships (NPR) to generalize effectively to previously unseen GAN sources.
This success can be attributed to NPR's unique ability to capture and analyze the distinctive traces left by the up-sampling component within common GAN pipelines. The insights gained from this localized analysis contribute to NPR's effectiveness across a spectrum of GAN-generated images.

%These results on 17 generation models indicate that the proposed artifacts representation from up-sampling operations is able to generalize to unseen GAN sources, while training on the ProGAN. 
%This can be attributed to NPR's capability of exploring the trace of up-sampling component in common GAN pipelines.

%\noindent\textbf{Diffusion-sources evaluation.}
\subsubsection{Diffusion-Sources Evaluation}
To present a more challenging evaluation scenario, we devise a comprehensive experiment where the detector is trained on images generated by ProGAN and subsequently tested on images produced by a diverse array of diffusion models. 
Three diffusion datasets are employed to preform evaluation on the diffusion-sources. 
It's important to note that the detectors are trained using the ProGAN 4-classes setting to ensure consistency.
This evaluation setup aims to assess the detector's adaptability and performance when faced with the inherent challenges posed by diffusion-generated images, providing valuable insights into the generalization capability of Neighboring Pixel Relationships (NPR) across different image generation techniques.

%To expand the testing scope, we further design a more challenging scenario,  training the detector with images generated by ProGAN then testing it on images produced by various diffusion models. Three diffusion datasets are employed to preform evaluation on the diffusion-sources, including ADM \cite{dhariwal2021diffusion}, DDPM \cite{ho2020denoising}, IDDPM \cite{nichol2021improved}, LDM \cite{rombach2022high}, PNDM \cite{liu2022pseudo}, Vqdiffusion \cite{gu2022vector}, Stable Diffusion v1 \cite{rombach2022high}, Stable Diffusion v2 \cite{rombach2022high},
%Midjourney, DALL-E \cite{ramesh2021zero}, Glide \cite{nichol2021glide}. The detectors still is trained with the ProGAN 4-classes setting.

The detection performance on DiffusionForensics \cite{Wang_2023_ICCV} is presented in Table \ref{tab:iccvdataset}. 
Despite being trained on images generated by ProGAN, NPR exhibits strong generalization capabilities across various diffusion models. Our method achieves 95.3\% and 99.8\% in terms of mean Accuracy (Acc.) and mean Average Precision (A.P.), respectively. NPR outperforms the current state-of-the-art methods LGrad and Ojha \cite{ojha2023towards} by 7.1\% and 20.9\%, respectively, in terms of mean Acc. metric. Additionally, when compared to DIRE \cite{Wang_2023_ICCV}, which specifically focuses on detection in the diffusion domain, NPR demonstrates comparable results, particularly noteworthy considering our training set comprises ProGAN images while DIRE relies on diffusion models for training.

%The detection performance on DiffusionForensics \cite{Wang_2023_ICCV} are shown in Table \ref{tab:iccvdataset}. 
%Despite training on images of ProGAN, NPR still keeps a strong generalization capability on the diffusion models. Our method achieves  95.3\%, 99.8\% in terms of mean Acc. and mean A.P., respectively. 
%Our NPR outperforms the current state-of-the-art LGrad and Ojha \cite{ojha2023towards}, by 7.1\% and 20.9\% in terms of mean Acc. metric. 
%In addition, compare to the DIRE \cite{Wang_2023_ICCV} focussing on detection in diffusion domain, our NPR obtains comparable results, when our training set is ProGAN and their training setting rely on diffusion models.

Given that a significant portion of images in DiffusionForensics \cite{Wang_2023_ICCV} belongs to the bedroom class, we further evaluate the performance on the diffusion dataset from Ojha \cite{ojha2023towards}. 
 In this dataset, diffusion models are utilized to generate images with 100 or 200 steps. The results on the Ojha diffusion dataset are presented in Table \ref{tab:cvprdataset}. The proposed NPR achieves a mean Accuracy (Acc.) value of 95.2\%, demonstrating its robust performance. When compared to the current state-of-the-art methods LGrad and Ojha \cite{ojha2023towards}, our NPR exhibits substantial improvements, surpassing these methods by 4.3\% and 8.3\% in mean accuracy. This comparison underscores NPR's ability to maintain satisfactory generalization across unseen diffusion datasets.

%Since most of images in the DiffusionForensics \cite{Wang_2023_ICCV} are bedroom class, we further perform evaluation on the diffusion dataset from Ojha \cite{ojha2023towards}. They perform 100 or  200 steps to generate image using diffusion models. 
%The results on  diffusion dataset from Ojha \cite{ojha2023towards} are shown in Table \ref{tab:cvprdataset}. The proposed NPR achieves a mean Acc. value of 95.2\%, demonstrating its strong performance.  In comparison to the current state-of-the-art methods LGrad and Ojha, our NPR exhibits substantial improvements, surpassing these methods by 4.3\% and 8.3\% in mean accuracy. 
% The comparison indicates that NPR maintains a satisfactory generalization capability even though facing unseen diffusion datasets.

The diffusion dataset from Ojha \cite{ojha2023towards} adopts only 100 or 200 steps to generate images, which may lack clarity and realism. To address this limitation, we further collect images of DALLE and Midjourney from the official Discord channels and sample images from other models with 1000 diffusion steps. The results are reported in Table \ref{tab:mydiffusion}. The NPR achieves gains of 15.9\% and 13.7\% compared to LGrad and Ojha, obtaining a mean accuracy value of 80.1\%. This result suggests that NPR maintains strong generalization performance even when faced with diffusion datasets generated with a more extended diffusion process (1000 steps).

%
%The diffusion dataset from Ojha \cite{ojha2023towards} adopt only 100 or  200 steps to generate image, lacking of clarity and reality. 
%Thus, we further collect images of DALLE and Midjourney from the official channel of Discord and sample images from other models with 1000 diffusion steps. The results are reported in Table \ref{tab:mydiffusion}. 
% The NPR achieves gains of  5.6\% and 14.4\% compared to LGrad and  Ojha, getting mean accuracy value of  86.4\%. 

In terms of the mean accuracy across 28 generation models, our NPR achieves 92.2\% shown in Table \ref{tab:all}, outperforming Ojha and LGrad by 12.4\% and 11.7\%, respectively. This result demonstrates that the proposed local up-sampling artifact, Neighboring Pixel Relationships, is capable of generalizing to both unseen GAN sources and diffusion sources, even when trained on ProGAN. This success can be attributed to the NPR's ability to rethink generator architectures and explore the trace of up-sampling from the perspective of local spatial information.

%In term of  mean Acc. of 28 generation models, our NPR achieves \ccn{**\%}, outperforming Ojha and LGrad \ccn{*\%}. 
%It  demonstrates that the proposed local un-sampling artifact, Neighboring Pixel Relationships, is able to generalize to both unseen GAN sources and diffusion sources, even training on the ProGAN.
%This can be attributed to the NPR rethink the generator architectures and explore the trace of up-sampling from the view of  local spatial information.

%\noindent\textbf{Performance on Face data.}
%
%
%
%\noindent\textbf{Compare to Edge-based representation}

\noindent\textbf{Different Upsampling Techniques.}
The performance of Neighboring Pixel Relationships on 28 generation techniques indicates a strong generalization ability to unseen sources. Despite being trained on ProGAN using nearest-neighbor up-sampling, the detector performs well on generation models with other up-sampling operations, such as bilinear. This phenomenon can be attributed to several factors:
1) Up-sampling operations are applied to feature maps, while NPR is exploited in image space.
2) NPR captures implicit artifacts representations caused by up-sampling operations.
3) The proposed NPR focuses on local and relative information, which enhances generalization ability across different upsampling techniques.

%The performance of Neighboring Pixel Relationships on 28 generation techniques indicate the generalization ability on unseen sources. 
%There are different upsampling operations in those generation architectures. 
%Although our detector trained on ProGAN using nearest neighbor upsampling, we  perform detection well on the generation models with other upsampling operations, such as bilinear. 
%The reason of  this phenomenon is that 
%1) the upsampling operations are applied on feature map, while the NPR is exploited in image space; 
%2) It is implicit representation for the upsampling artifact caused by upsampling operations; 
%3) the proposed NPR focus on the local information, which is benefit to enhance the generalization ability.

\noindent\textbf{Effect of choice of NPR‘s hyperparameters.}
We evaluate the impact of NPR's size $l$ and index $j$ in Equation \ref{eq:eq4} on generalization ability. Simultaneously, to validate the effectiveness of the subtraction between elements in Equation \ref{eq:eq4}, we replace $w_j$ with $avg(v_I^c)$ and $max(v_I^c)$ to implement NPR. We employ the $(car, cat, chair, horse)$ of ProGAN as the training set and apply the test set of ForenSynths\cite{wang2020cnn} for evaluation. The results are shown in Table \ref{tab:tab6}. Observations:
1) When $l=2$, NPR achieves better performance, likely due to most generators employing 2 scaled up-sampling layers.
%2) NPR with $max(v_I^c)$ does not generalize well to other GANs.
2) NPR with $avg(v_I^c)$ and $max(v_I^c)$ show similar detection performance.
This suggests that information in the $2 \times 2$ block of images can effectively reveal differences between real and fake images.

%
%We evaluate the impact of NPR's size $l$ and index $j$ in the Equation \ref{eq:eq4} on the generalization ability.
%Meanwhile, to valid the effectiveness of the Subtraction between elements in the Equation \ref{eq:eq4},  we also replace the $w_j$ to $avg(v_I^c)$ and $max(v_I^c)$ to implement the NPR. We employ the $(car, cat, chair, horse)$ of ProGAN as the training set, and apply test set of ForenSynths\cite{wang2020cnn} for evaluation. The results are shown in Table \ref{tab:tab6}.
%We can observe that 1)  when $l=2$, the NPR can obtain better performance. It is caused that most of generators employ 2 scaled up-sampling layer. 2) the NPR with $max(v_I^c)$ is not able to generalize well to other GANs. 3) the NPR with $avg(v_I^c)$ get similar detection performance. 
%This  suggesting that the information in the  $2 \times 2$ block of images can effectively reveal the differences between real and fake images.

 \begin{figure}[ht!]
%\vspace{-0.25 cm}
   \centering
    \includegraphics[scale=0.280]{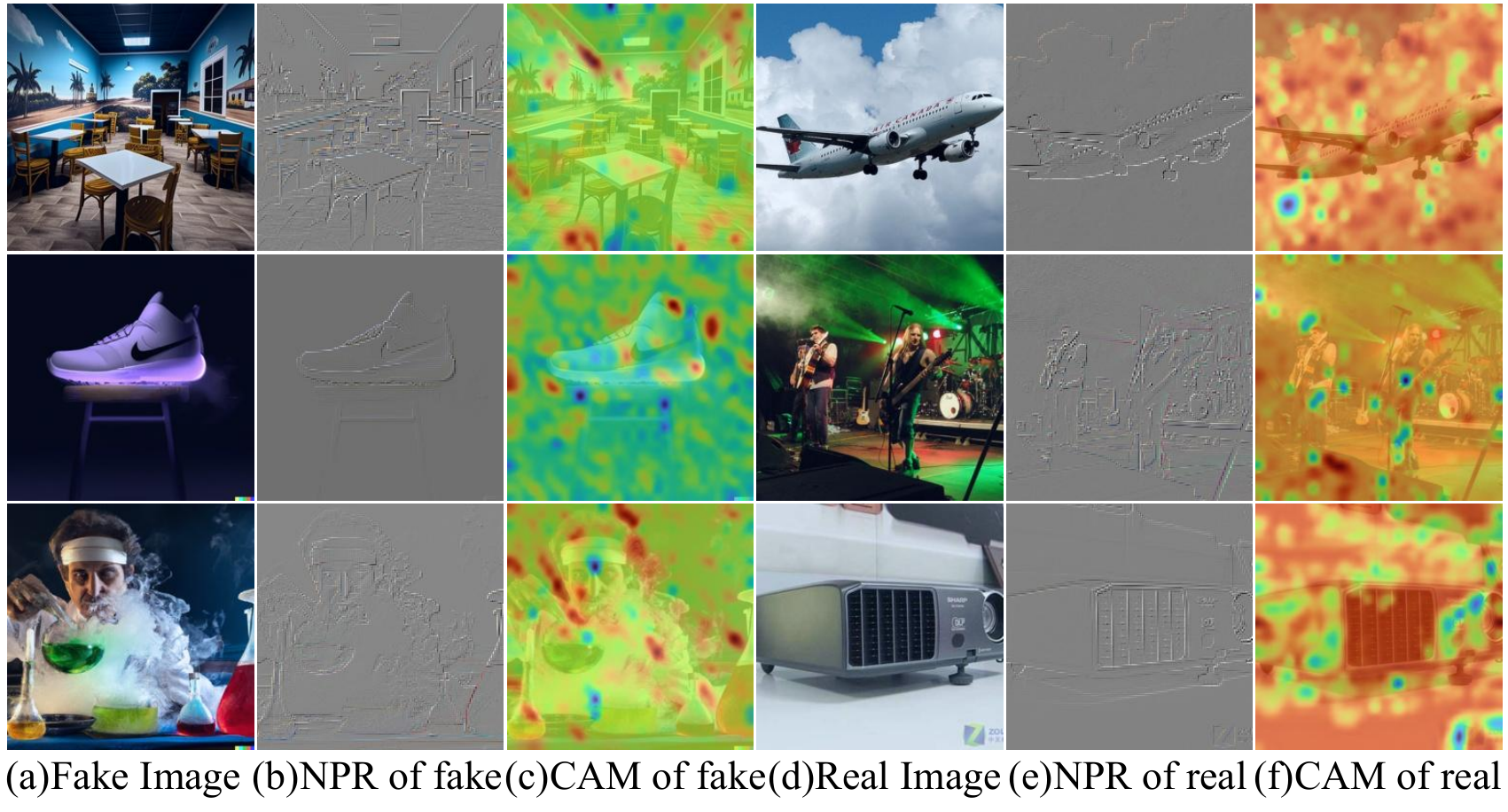}
    \caption{The visualization of CAM \cite{zhou2016learning} extracted from detector on image of Midjourney, DALLE, and ImageNet. Warmer color indicates a higher
probability. }
    \label{fig:cam}
    \vspace{-0.15 cm}
 \end{figure}

%\vspace{-0.25 cm}
\noindent\textbf{Qualitative Analysis of NPR.} 
%The key innovation of our NPR lies in present a simple and generalized artifact representation from the common up-sampling component of generation pipelines.
The above quantitative experiments have indicated the effectiveness of the proposed Neighboring Pixel Relationships. 
To obtain a more profound understanding of its intrinsic properties, we conduct a qualitative analysis of NPR, employing the visualization of Class Activation Map (CAM) \cite{zhou2016learning}. Figure \ref{fig:cam} illustrates the Class Activation Maps for images sourced from Midjourney, DALLE, and ImageNet. Notably, the CAMs for real images highlight a broader portion of the image, whereas the CAMs for fake images tend to emphasize localized regions.Intriguingly, despite the detector being primarily trained on a dataset encompassing cars, cats, chairs, and horses, it demonstrates the capacity to recognize these diverse images.
Certainly, this emphasizes the generalization ability of our detector in identifying various deepfake signatures, showcasing its capacity to extend recognition capabilities beyond the training classes.

%To gain a better understanding of its intrinsic properties, we conduct a further qualitative analysis of NPR, utilizing visualization of Class Activate Map \cite{zhou2016learning}. 
%The Fig. \ref{fig:cam} shows the Class Activate Map of images from Midjourney, DALLE, and ImageNet. It's noteworthy that the CAMs for real images highlight a broader portion of the image, while the CAMs for fake images tend to emphasize localized regions.Interestingly, even though the detector is
%primarily trained using a dataset containing cars, cats, chairs,
%and horses, it showcases the ability to recognize those images.

% It is evident that real images highlight the face region while fake images emphasize the background region. Although the detector is trained using sets of cars, cats, chairs, and horses, it can still recognize face images.

%\noindent\textbf{Limitation.}

\section{Conclusion}

This work focuses on developing a generalizable artifacts representation for both GANs and diffusions detection. 
We reconsider the architectures of CNN-based generators, aiming to establish source-invariant forgery detection. 
Our findings reveal that the up-sampling operator, beyond frequency-based artifacts, can produce generalized forgery artifacts. 
Existing works typically consider its influence on the whole image in the frequency domain. In contrast, we explore the trace of the up-sampling layer from the local image pixels. 
We present a simple but effective artifact representation, named Neighboring Pixel Relationships (NPR), to achieve generalized deepfake detection. 
Extensive experiments on 28 generation models indicate that the proposed representation NPR contributes to a strong AI-generated image detector.

%
%This works focus on developing a generalizable artifacts representation for both GANs and  diffusions detection. 
%We rethink the architectures of CNN-based generator, establishing the source-invariant forgery detection. 
%Our findings illuminate that the up-sampling operator can, beyond frequency-based artifacts, produce generalized forgery artifacts. 
%Existing works only consider its influence on the whole image in the frequency domain. In contrast, we explore the trace of up-sampling layer from the local image pixels.
%We present a simple but effective artifact representation, named Neighboring Pixel Relationships (NPR), to achieve generalized deepfake detection. 
%Extensive experiments on 28 generation models indicate that the proposed representation NPR contributes to a strong AI-generated image detector.

%\input{sec/3_finalcopy}
{
    \small
    \bibliographystyle{ieeenat_fullname}
    \bibliography{NPR}
}


\end{document}